\documentclass[runningheads]{llncs}
\usepackage[T1]{fontenc}
\usepackage{graphicx}
\usepackage{amsmath, amssymb, mathtools}
\usepackage{hyperref}
\usepackage{cleveref}
\usepackage{algorithm}
\usepackage{algorithmic}
\usepackage{booktabs}
\usepackage{multirow}
\usepackage{xcolor}
\usepackage{subcaption}
\usepackage{adjustbox}

\usepackage{xspace}

\newcommand{\x}{\mathbf{x}}
\newcommand{\xt}{\mathbf{x}^t}

\newcommand{\CM}[2]{%
  \ifnum#1=1
  \else
    #2
  \fi
}

\usepackage{bbm}

\newcommand{\defeq}{\vcentcolon=}

\newcommand{\norm}[1]{\left\| #1 \right\|}

\DeclareMathOperator*{\argmin}{arg\,min}

\newcommand{\avgn}{\frac{1}{n} \sum_{i=1}^{n}}
\newcommand{\avgcn}[1]{\frac{#1}{n} \sum_{i=1}^{n}}
\newcommand{\avgcnn}[1]{\frac{#1}{2n} \sum_{i=1}^{n}}

\newcommand{\lrp}[2][]{%
  \if\relax\detokenize{#1}\relax%
    \left(#2\right)%
  \else%
    {\color{#1}\left(\right.}#2{\color{#1}\left.\right)}%
  \fi%
}

\newcommand{\R}{\mathbb{R}}
\newcommand{\E}{\mathbb{E}}

\newcommand{\bx}{\mathbf{x}}

\newcommand{\bxt}{\mathbf{x}^t}
\newcommand{\bxtt}{\mathbf{x}^{t+1}}

\newcommand{\bxit}{\mathbf{x}_i^t}

\newcommand{\bxstar}{\mathbf{x}^\star}

\newcommand{\bxlambdastar}{\mathbf{x}_\lambda^\star}
\newcommand{\bxgstar}{\mathbf{x}_\gamma^\star}
\newcommand{\bxqstar}{\mathbf{x}_q^\star}

\newcommand{\abar}{\bar{a}}

\newcommand{\bg}{\mathbf{g}}
\newcommand{\bgbar}{\bar{\mathbf{g}}}

\newcommand{\gradfi}{\nabla f_i}
\newcommand{\gradF}{\nabla F}

\usepackage{comment}

\providecommand{\inlinecomment}[3]{%

  {\color{#1}#2: #3}}%
\newcommand\commenter[2]%
{%
  \expandafter\newcommand\csname i#1\endcsname[1]{\inlinecomment{#2}{#1}{##1}}
  \expandafter\newcommand\csname #1\endcsname[1]{\mycomment{#1}{##1}{#2}}
}

\newcommand{\cm}[1]{}

\definecolor{electricviolet}{rgb}{0.56, 0.0, 1.0}

\begin{document}

\title{FAIRVAR: Fair Federated Learning via Variance Regularization}
\author{
Zahra Kharaghani\inst{1}
\and
Ali Dadras\inst{2}
\and
Tommy Löfstedt\inst{1}
}

\authorrunning{Z. Kharaghani et al.}

\institute{
Department of Computing Science, Umeå University, Umeå, Sweden
\email{zahra.kharaghani@slu.se,}
\email{tommy.lofstedt@umu.se}
\and
Department of Mathematics and Mathematical Statistics, Umeå University, Umeå, Sweden
\email{ali.dadras@umu.se}
}
\maketitle            

\begin{abstract}
Federated learning (FL) allows collaborative training of machine learning models across multiple parties without sharing raw data. However, heterogeneous data can cause some clients to have disproportionate influence on the global model, leading to disparities in their performance. Fairness, understood as reducing these disparities, is therefore a crucial concern in FL and has been addressed in various ways.
We studied performance equitable fairness in FL, where the goal is to minimize performance disparities across clients.
We evaluated several existing fairness-aware methods and introduce here a new gradient-variance-regularized method, implemented in two variants: FairGrad (approximate) and FairGrad$^*$ (exact).
We theoretically characterize the connections between these methods and, empirically, on heterogeneous benchmarks, show that FairGrad and FairGrad$^*$ consistently improve fairness by reducing variance in client accuracies, while maintaining competitive or improved mean performance compared to existing fairness-aware baselines.

\end{abstract}

\section{Introduction}
\label{sec:Introduction}
% Introduction
In federated learning (FL), it is often assumed that individual clients have insufficient data to effectively train machine learning models on their local datasets. By collaborating within the FL framework, clients can jointly train a global model, and by that enhance overall performance without sharing raw data.
However, due to data heterogeneity, the global model may perform inconsistently across clients. This disparity poses a significant challenge, particularly for clients with substantially different data distributions or smaller sample sizes. Such clients may experience poor performance when using the global model.
To address this issue, it is crucial to ensure that the training process in FL is fair, \emph{i.e.}, for clients to have an incentive to participate, they must benefit from the global model without being disadvantaged by disparities in their local data distributions.

Fairness in FL is generally categorized into three types:  collaborative fairness, equitable fairness, and algorithmic fairness. 
In this work, we focused on equitable fairness, which seeks to minimize disparities in client performances. Within this category, fairness has been further conceptualized in three distinct forms:
good-intent equitable fairness~\cite{mohri2019agnostic}, Pareto-optimal equitable fairness~\cite{cui2021addressing}, and performance equitable fairness~\cite{li2019fair}.
Among these research lines, performance equitable fairness has attracted much attention. This approach aims to ensure equal performance across all clients without sacrificing the overall model performance. The local loss functions are usually modified in an attempt to achieve performance equitable fairness.
It is, however, still unclear how different proposed loss functions support the goal of performance equitable fairness, making it challenging to determine which approaches are most effective in practical settings. Moreover, most existing methods lack a solid theoretical foundation, making it difficult to justify their use or apply them across diverse FL scenarios.

In this work, we analyzed and evaluated several methods for performance equitable fairness in FL. We distinguished between existing fairness-aware methods, baselines, and our proposed contribution. The existing fairness-aware methods include loss-variance regularization~\cite{ling2024fedfdp} and q-Fair Federated Learning
(q-FFL;~\cite{li2019fair}), which, as we show later, is related to loss-variance regularization.
We proposed a
new gradient-variance regularization approach for performance equitable fairness,
implemented in two variants: FairGrad, an approximate and communication-efficient version, and FairGrad$^*$, an exact version. 
As baselines, we also included federated averaging (FedAvg;~\cite{mcmahan2017communication}), Agnostic Federated Learning (AFL;~\cite{mohri2019agnostic}), and the more recent AAggFF~\cite{hahn2024pursuing}.
Formal definitions of the variance-regularized loss functions are provided in \Cref{sec:Statement_of_the_Problem}.

Our analysis revealed how these objectives are related under homogeneous and heterogeneous settings. \Cref{sec:Analysis of Stationary Points} provides theoretical insights into the relationships between FedAvg, q-FFL, and two variance-regularized methods, and shows how variance regularization can change the optimization behavior under heterogeneity. Empirical results further demonstrated that the variance-regularized methods generally outperform the baselines in both fairness and overall performance, with the proposed variants, FairGrad and FairGrad$^*$, showing particularly strong results. The code is available at: \url{https://github.com/zahrakhara/FAIRVAR}.

\section{Related Work}
\label{sec:Related work}
% Related Work

Fairness in FL can be defined in different ways: collaborative fairness~\cite{sim2020collaborative,xu2020reputation,rashid2025trustworthy}, which allocates rewards proportionally to clients' data contributions, 
algorithmic fairness~\cite{cui2021addressing,ezzeldin2023fairfed,su2024multi,chen2025afed,he2025towards}, which ensures that the trained model is not biased against any client, and equitable fairness, which minimizes
performance disparities across clients. 
Equitable fairness includes good-intent  fairness~\cite{mohri2019agnostic}, performance equitable fairness~\cite{li2019fair}, and Pareto-optimal fairness~\cite{cui2021addressing}.
Equitable fairness was introduced in~\cite{mohri2019agnostic}, where AFL was proposed. They showed that AFL naturally leads to a notion of fairness they refer to as good-intent fairness. This method aims to find the best global model by focusing on the worst-case client performance. However, significant performance disparities can still occur under this approach~\cite{nguyen2024federated}, and AFL has been shown to work best with smaller numbers of clients~\cite{li2019fair}.
In contrast to AFL, q-FFL~\cite{li2019fair} offers greater flexibility 
and can be fine-tuned to achieve the desired level of fairness.  
q-FFL utilizes a method called q-FedAvg, which focuses on fairness by dynamically re-weighting clients based on their performance. Clients with lower performance are given higher weights to ensure a more uniform performance distribution across clients. 
The fairness level is controlled by the parameter $q$, with larger values placing more emphasis on disadvantaged clients.
AFL and FedAvg are special cases of q-FFL, obtained by setting $q$ to a sufficiently large value and $q=0$, respectively.

AAggFF~\cite{hahn2024pursuing} builds on this line of work by unifying existing fairness-aware aggregation strategies within an online convex optimization framework and enhancing them through sequential decision-making at the server, with variants for both cross-device and cross-silo FL.
Another direction studies fairness in FL via bilevel optimization. FedFB~\cite{zeng2021improving} selects fair batch representations, while entropy-based aggregation~\cite{wang2024entropy} aligns models and gradients through adaptive updates.
Some works highlight that data heterogeneity can lead to different levels of irreducible error across clients. Pareto-optimal fairness~\cite{cui2021addressing} addresses this by minimizing worst-case loss while preserving overall performance. Building on this idea, methods such as FedMGDA+~\cite{hu2022federated} and AdaFed~\cite{hamidi2024adafed} extend multi-objective optimization for fair FL.

Variance-regularized methods have been explored in FL for different purposes. For instance, \cite{guo2023out} proposed FEDIIR, which uses gradient variance regularization to enhance out-of-distribution generalization for non-participating clients. In contrast, we propose an approach
that applies gradient variance regularization specifically to improve performance equitable fairness among participating clients in heterogeneous settings, supported by both theoretical analysis and empirical evaluation.

While various directions have been explored, the present work focused on performance equitable fairness, where important gaps remain: The theoretical connections among existing approaches are not well understood, and their performance under heterogeneous data has not been systematically evaluated. In this work, we addressed these gaps and clarified some of the theoretical connections, conceived new variance-regularized methods, and evaluated their effectiveness under heterogeneous data (Sections \ref{sec:Analysis of Stationary Points}--\ref{sec:Experiments}).

\section{Statement of the Problem}
\label{sec:Statement_of_the_Problem}
% Statement of the Problem
The standard FL procedure, 
aims to minimize a global average expected risk,
\begin{equation}\label{sec:Problem_Statement:Fedavg_p}
    \min_{\bx \in \R^d} \left\{
        \widetilde{F}(\bx) \defeq \avgn \tilde{f}_i(\bx) =  \avgn  \mathbb{E}_{\xi_i \sim \mathcal{D}_i} \big[\ell(\bx, \xi_i)\big]
    \right\},
\end{equation}
where $n$ denotes the number of participants (clients) involved in the federation, $\bx \in \mathbb{R}^d$ is the parameters of the machine learning model, $\xi_i\sim\mathcal{D}_i$ is a sample data point from the $i$-th client's data distribution, $\mathcal{D}_i$, and $\ell(\bx, \xi_i)$ is the loss function, indicating the loss of the model with parameters $\bx$ on data sample $\xi_i$. As the true data distribution, $\mathcal{D}_i$, is unknown, clients rely on an empirical set of $N_i$ samples, $\mathcal{S}_i=\{\xi_i^1, \ldots, \xi_i^{N_i}\} \sim\mathcal{D}_i^{N_i}$, and corresponding global empirical risk,
\begin{equation}\label{sec:Problem_Statement:Fedavg}
    \min_{\bx \in \R^d} \Bigg\{
        F(\bx)
            \defeq \avgn {f}_i(\bx)
            =  \avgn\frac{1}{N_i }\sum_{j=1}^{N_i} \ell(\mathbf{x}, \xi_i^j)
    \Bigg\}.
\end{equation}
In the rest of the paper, $F(\bx)$ denotes the global empirical risk, also referred to as the average loss across clients.
Solutions to \Cref{sec:Problem_Statement:Fedavg} may result in significant performance discrepancies among participating clients, particularly when the number of samples, $N_i$, is small or the client data distributions are highly heterogeneous~\cite{zhao2018federated}. To quantify these discrepancies, we outline a fairness criterion for FL, building on but slightly modifying the definitions introduced by~\cite{li2019fair} and~\cite{nguyen2024federated}.
\begin{definition}[Performance equitable fairness in FL]
    \label{definition_fairness}
    For a set of models, with weights $\mathcal{X} = \{ \mathbf{x} : |F(\mathbf{x}) - \min_{\mathbf{x}'} F(\mathbf{x}')| \leq \varepsilon \}$, where $\varepsilon\geq0$ is a tolerance for performance degradation, a model $\mathbf{x}_1 \in \mathcal{X}$ is said to achieve better $\varepsilon$-performance equitable fairness than model $\mathbf{x}_2 \in \mathcal{X}$ if the variance of its performance across clients is lower. That is, if $\text{var}((p_i(\mathbf{x}_1))_{i=1}^n) < \text{var}((p_i(\mathbf{x}_2))_{i=1}^n)$, where
    $p_i(\mathbf{x})$ denotes a performance measure (\textit{e.g.}, accuracy) for model $\mathbf{x}$ on the $i$-th client's local data and $\text{var}((p_i(\mathbf{x}))_{i=1}^n)$ denotes the sample variance across all clients.
\end{definition}
When a model is trained on data from multiple clients, fairness per \Cref{definition_fairness} means that each client has an equal opportunity to benefit from the model. The goal is to achieve equitable performance across all clients, regardless of differences in the data they contribute with. The tolerance parameter, $\varepsilon$, ensures that only models with average loss, $F(\mathbf{x})$, close to the best model (within $\varepsilon$) are considered. This avoids favoring models that achieve low variance simply because they perform poorly on all clients. Instead, the comparison is restricted to high-performing models, among which the fairest model is preferred.

\subsection{Evaluated Methods}
In this section, we present the objective functions of the methods evaluated in this work for performance equitable fairness in FL. These methods aim to reduce disparities in client performance by addressing differences in client losses or gradients. Each method introduces a tunable parameter that controls the trade-off between average performance and fairness. When set to zero, the objective reduces to the standard finite-sum formulation in Equation~\eqref{sec:Problem_Statement:Fedavg}; otherwise, it promotes a more uniform distribution of performance across clients, potentially improving outcomes for under-performing clients at the expense of others.

Ling \textit{et al.}~\cite{ling2024fedfdp} proposed to penalize models with high variance in the client performances, which can be implemented using a variance-regularized loss, as
\begin{equation}\label{eq:fairFL}
    \min_{\bx \in \R^d} \Bigg\{
        L_{\lambda}(\bx)
        \defeq F(\bx) 
            + \avgcnn{\lambda} \big(f_i(\bx) - F(\bx) \big)^2
    \Bigg\},
\end{equation}
where $\lambda$ controls the degree of fairness and $F(\bx)$ denotes the average loss, as defined in Equation~\eqref{sec:Problem_Statement:Fedavg}. We propose two new algorithms to solve this objective, detailed in Algorithm~\ref{app:algo:loss} and Algorithm~\ref{algo:exact:lossreall}.

We further propose a novel regularization approach aiming to promote performance equity across participating clients in FL by minimizing gradient disparities. This proposed method is formulated as 
\begin{equation}\label{eq:fairGrad}
    \min_{\bx \in \R^d} \Bigg\{
        J_{\gamma}(\bx)
        \defeq F(\bx) 
            + \avgcnn{\gamma} \big\|\nabla f_i(\bx) - \nabla F(\bx) \big\|^2
    \Bigg\},
\end{equation}
where $\nabla F(\bx)\defeq\avgn \nabla f_i(\bx)$ denotes the average gradient, $\gamma$ controls the degree of fairness, $\|\cdot\|$ denotes the Euclidean norm, and $F(\bx)$ is defined in Equation~\eqref{sec:Problem_Statement:Fedavg}, as before. We also propose two new algorithms to solve \Cref{eq:fairGrad}, that are detailed in Algorithm~\ref{sec:app:algo:grad} and Algorithm~\ref{sec:app:algo:gradreall}.

Li \textit{et al.}~\cite{li2019fair} proposed q-FFL to improve fairness in FL by prioritizing clients with higher losses through a reweighting mechanism that gives higher weight to under-performing clients. It is formulated as
\begin{equation}\label{qffl}
    \min_{\bx \in \R^d}\Bigg\{
        H_q(\bx)
            \defeq \sum_{i=1}^{n} \frac{r_i}{q + 1} \big( f_i(\bx) \big)^{q + 1}
    \Bigg\},
\end{equation}
where $r_i$ is the proportion of total data samples held by client $i$, and 
$q$ is a parameter that controls the degree of fairness.

The FedAvg algorithm~\cite{mcmahan2017communication} aims to minimize the average of local loss functions across clients. The objective, given in~\Cref{sec:Problem_Statement:Fedavg}, is considered here with uniform weights, $1/n$, for all clients.
Equations~\eqref{eq:fairFL}, \eqref{eq:fairGrad}, and \eqref{qffl}  define the objective functions of the fairness methods considered in this work, however, they are not directly solvable in a federated context because they are not separable with respect to each client's objective. To address this, we propose both exact and approximate algorithms for Equations~\eqref{eq:fairFL} and~\eqref{eq:fairGrad} in \Cref{sec:Algorithms}, while for q-FFL we adopt the original optimization algorithm proposed by Li \textit{et al.}~\cite{li2019fair}.

\section{Analysis of Stationary Points}
\label{sec:Analysis of Stationary Points}
% Analysis of Stationary Points
All proofs for the theoretical results in this section are provided in \Cref{appendix}.
\textbf{Homogeneous Data}:
In a homogeneous data environment, the datasets of all $n$ clients are drawn from the same underlying probability distribution, \emph{i.e.}, $\mathcal{D}_i = \mathcal{D}_j$ for all $i, j\in\{1,\ldots,n\}$.
As a result, each client's local loss, $f_i(\bx)$, is an unbiased estimator of the expected average loss, $\E[F(\bx)]$, and each client's gradient, $\nabla f_i(\bx)$, is an unbiased estimator of the expected average gradient, $\E[\nabla F(\bx)]$.
This implies that, in homogeneous settings, introducing the regularization terms in Equations~\eqref{eq:fairFL} and \eqref{eq:fairGrad} does not alter the location of the stationary points on average, thereby preserving the optimal solutions of the original objective, $F(\bx)$.
Consequently, the regularization terms are unnecessary in such settings, as they do not change the outcome.
This conclusion also holds for q-FFL, as the following theorem implies. We show that the expectation of the function $H_q(\bx)$, when $q=1$, can be upper bounded by the expectation of the function $L_{\lambda}(\bx)$, for a particular value of $\lambda$. 
\label{qffl-relation-loss}
\newcommand{\qfflrelationlosstext}{
    For $q=1$ and $r_i=\frac{1}{n}$, we have 
    \begin{align}
    \E\big[H_1(\bx)\big]
       = \frac{1}{2n}\sum_{i=1}^n \E\big[f_i(\bx)\big]^2 \nonumber 
        + \frac{n}{n-1}\E\!\left[
            \frac{1}{2n}\sum_{i=1}^n \big(f_i(\bx) - F(\bx)\big)^2
         \right]. \nonumber
    \end{align}
    If $\E[f_i(\bx)] \leq C$ for all $i$ and some constant $C$, then
     $\E[H_1(\bx)] \leq \E[L_{\lambda}(\bx)]$,
    with $\lambda=\frac{n}{C(n-1)}$.}
\begin{theorem}
\label{qffl-relation-loss}
\qfflrelationlosstext
\end{theorem}
\Cref{qffl-relation-loss} demonstrates that $H_1(\bx)$, \emph{i.e.}, q-FFL with $q=1$, can be expressed in a similar form as $L_\lambda(\bx)$ for a particular $\lambda$. Further, at the stationary points of $F(\bx)$, the two functions $H_q(\bx)$ and $L_{\lambda}(\bx)$ (for $q=1$ and $\lambda=\frac{n}{C(n-1)}$) are \textit{close}, as $F(\bx)$ attains its minimum value at these points, leading to a small difference between them.

\noindent\textbf{Heterogeneous Data}:
In the case of heterogeneous data, the datasets of different clients are drawn from different distributions, \emph{i.e.}, $\mathcal{D}_i \neq \mathcal{D}_j$ for some $i, j\in\{1,\ldots,n\}$, resulting in variability in the local objectives and gradients. In this case, the variance terms in the regularized loss functions, $L_{\lambda}(\bx)$, $J_\gamma(\bx)$, and $H_q(\bx)$ with $q=1$, are important because they penalize this variability. The regularization encourages the optimization process to find solutions that balance performance across diverse clients, promoting models that generalize better.
While the expectation-based arguments in the homogeneous case rely on the unbiasedness of local losses, and local gradients with respect to the global objective, 
such assumptions do not hold universally in the heterogeneous setting, where client data distributions differ. However, under certain conditions, specifically when client losses or gradients are aligned at the optimum point, similar results can still be obtained.
\newcommand{\stationarypointthmthreetext}{
    Assume that the $f_i$'s are differentiable. Let $\bxstar \in \argmin_{\bx \in \R^d} F(\bx)$, $\bxlambdastar \in \argmin_{\bx \in \R^d} L_{\lambda}(\bx)$, $\bxgstar \in \argmin_{\bx \in \R^d}J_{\gamma}(\bx)$, and $\bxqstar \in \argmin_{\bx \in \R^d}H_q(\bx)$,
    such that 
    \[
        \nabla F(\bxstar)=\nabla L_{\lambda}(\bxlambdastar)=\nabla J_\gamma(\bxgstar)=\nabla H_q(\bxqstar) = 0.
    \]
    If either $f_i(\bxstar)=f_j(\bxstar)$ or $\nabla f_i(\bxstar)=\nabla f_j(\bxstar)$ holds for all $i,j$, then we have $\nabla L_{\lambda}(\bxstar)=\nabla J_\gamma(\bxstar)=\nabla H_q(\bxstar) =0$.
    Similarly, if either $f_i(\bxlambdastar)=f_j(\bxlambdastar)$ or $\nabla f_i(\bxlambdastar)=\nabla f_j(\bxlambdastar)$ for all $i,j$, then $\nabla F(\bxlambdastar)=0$. If the analogous condition holds at $\bxgstar$, and $\bxqstar$,  then $\nabla F(\bxgstar)=0$, and $\nabla F(\bxqstar)=0$.}
\begin{theorem}
\label{stationary-point-thm-three}
\stationarypointthmthreetext
\end{theorem} 
The following theorem investigates when stationary points differ between $F(\bx)$ and $L_{\lambda}(\bx)$.
This displacement is useful when the stationary points of the $F(\bx)$ lead to unequal client performance, as can occur under heterogeneous data. In such cases, variance regularization can move the solution away from an average-loss optimum toward a point with more balanced performance across clients.
\newcommand{\stationarypointthmonetext}{
Assume that the $f_i$’s are differentiable. Let 
$\bxstar \in \argmin_{\bx \in \R^d} F(\bx)$ with $\nabla F(\bxstar)=0$,  
and $\bxlambdastar \in \argmin_{\bx \in \R^d} L_{\lambda}(\bx)$ with $\nabla L_{\lambda}(\bxlambdastar)=0$.  
Furthermore, suppose there exist at least two indices $1 \leq i,j \leq n$ such that 
$f_i(\bx) \neq f_j(\bx)$ and $\nabla f_i(\bx) \neq \nabla f_j(\bx)$ 
for $\bx \in \{\bxstar,\bxlambdastar\}$.  
Let $\mathcal{G}(\bx) = \{\nabla f_i(\bx)\}_{i=1}^{n}$ denote the set of gradient vectors at $\bx$.  
Then
$\nabla L_{\lambda}(\bxstar) \neq 0 
\quad \text{and} \quad 
\nabla F(\bxlambdastar) \neq 0$
if at least one of the following conditions holds:
\begin{enumerate}
    \item Subset of clients with distinct gradient directions, others with identical loss values. That is, there exist $m$ linearly independent gradients in $\mathcal{G}(\bx)$, where $1 < m \leq n-1$,  
    and the remaining $n-m$ gradients correspond to loss functions that are equal at  
    $\bx \in \{\bxstar,\bxlambdastar\}$.  
    Specifically, if $I_{\text{ind}} \subseteq \{1,\dots,n\}$ indexes the $m$ linearly independent gradients, i.e.  
    $\mathcal{G}_{\text{ind}}(\bx) = \{\nabla f_i(\bx) \mid i \in I_{\text{ind}}\} \subseteq \mathcal{G}(\bx)$,  
    and the remaining indices are $I_{\text{rem}} = \{1,\dots,n\} \setminus I_{\text{ind}}$,  
    then the corresponding loss values satisfy $f_j(\bx) = f_k(\bx)$ for all $j,k \in I_{\text{rem}}$.
    \item Subset of clients with non-zero gradients, all but one sharing the same loss value.
   That is, there exist $m$ non-zero gradient vectors with $1 \leq m \leq n$,  
    and the loss functions corresponding to these gradients are all equal except for one.
\end{enumerate}}
\begin{theorem}
\label{stationary-point-thm-one}
\stationarypointthmonetext
\end{theorem}
The conditions in \Cref{stationary-point-thm-one} describe when fairness regularization alters the optimization landscape, such that a stationary point of one objective ($\bxstar$ for $F(\bx)$ or $\bxlambdastar$ for $L_{\lambda}(\bx)$) is not preserved for the other. Such situations arise naturally in heterogeneous FL.
The first condition occurs when clients have substantially different data distributions, leading to distinct gradient directions even when some clients exhibit similar loss values. The second arises when most clients have similar performance, but one client’s data distribution results in a noticeably different loss. In both scenarios, the disagreement in either gradient directions or loss values introduces an imbalance that fairness regularization alters, thereby shifting the stationary point between $F(\bx)$ and $L_{\lambda}(\bx)$. 

The following theorem shows that even when all clients have identical losses at a stationary point of $F(\bx)$, small zero-mean perturbations to the client losses can cause the stationary point of $L_{\lambda}(\bx)$ to differ from that of $F(\bx)$.
\newcommand{\stationarypointthmfourtext}{
   Assume that the $f_i$'s are differentiable. Let  $ \bxstar \in \argmin_{\bx \in \R^d} F(\bx)$ such that $f_i(\bxstar)=f_j(\bxstar)$ for all indices $i,j$.
    Assume further that, $\nabla f_i(\bxstar) \neq 0$ for at least one index $i$. Then, there exist non-zero values $o_i$ such that $\bar{o}:=\avgcn{1}o_i=0$ and the function $\hat{L}_\lambda(\bx)$ defined as
        $\hat{L}_{\lambda}(\bx):=\hat{F}(\bx) + \avgcnn{\lambda} \big(\hat{f}_i(\bx) - \hat{F}(\bx) \big)^2$,
    where $\hat{f}_i(\bx)=f_i(\bx) + o_i$, and $\hat{F}(\bx)=F(\bx)=\avgn  \hat{f}_i(\bx)$,
    has a different stationary point than $F(\bx)$.}
\begin{theorem}
\label{stationary-point-thm-four}
\stationarypointthmfourtext
\end{theorem}
We show that, under certain conditions, the regularized functions can converge to different stationary points of $F(\bx)$. The theorem in the appendix (\Cref{stationary-point-thm-five}) establishes that stationary points of $L_{\lambda}(\bx)$ can reduce the variance of local loss functions, leading to more balanced performance across clients. For brevity, the full statement and discussion are deferred to the \Cref{heterogeneous:data}.

\section{Algorithms}
\label{sec:Algorithms}
% Algorithms
To minimize the objective in \Cref{eq:fairFL}, which aims to minimize  both the average and the variance of client loss functions, we propose and evaluate two algorithms:
An approximate version (FairLoss; see Algorithm~\ref{app:algo:loss}) and an exact version (FairLoss$^*$; see  Algorithm~\ref{algo:exact:lossreall}).  
Similarly, for the objective in \Cref{eq:fairGrad}, which minimizes the average loss and reduces the variance of the gradients of
client loss functions, we introduce two algorithms:  
An approximate version (FairGrad; see Algorithm~\ref{sec:app:algo:grad}) and an exact version (FairGrad$^*$; see Algorithm~\ref{sec:app:algo:gradreall}).  
In the approximate variants, clients perform local updates using information from the previous iteration, thereby reducing communication with the server.  
In the exact variants, clients use information from the current iteration, which provides more accurate updates but at the cost of increased communication.  
Detailed descriptions of all algorithms are provided in \Cref{app:Algorithm:Descriptions}.    
In the following theorem, we show that the surrogate objective and gradient used by each approximate algorithm are consistent with those of its exact counterpart.
\newcommand{\approximatlossthmalgonetext}{
Let $\abar_{\lambda}(\bxt)$ and $\abar_{\gamma}(\bxt)$ denote approximations of the functions $L_{\lambda}(\xt)$ and $J_{\gamma}(\xt)$, respectively, and let $\bgbar_{\lambda}(\bxt)$ and $\bgbar_{\gamma}(\bxt)$ denote the corresponding gradient approximations at iteration $t$, as used in Algorithms~\ref{app:algo:loss}--\ref{sec:app:algo:grad}. We assume that SGD is used for the updates.
Then we have 
\[
    L_{\lambda}(\xt)- \abar_{\lambda}(\bxt) 
    = - \tfrac{\lambda}{2}\big(F(\bxt) - F(\bx^{t-1}) \big)^2 \leq 0, \text{~and}
\]
\[
    J_{\gamma}(\xt)- \abar_{\gamma}(\bxt) 
    \leq - \tfrac{\gamma}{2} \big\|\nabla F(\bxt) - \nabla F(\bx^{t-1})\big\|^2 \leq 0.
\]
Moreover, if each local loss function, $f_i(\bx)$, is $L$-smooth and $\eta$ denotes the learning rate, then $\|\nabla L_{\lambda}(\xt)\| 
        \leq \|\bgbar_{\lambda}(\bxt)\| + \lambda L \eta \|\bgbar_{\lambda}(\bx^{t-1})\|\, |F(\xt) - F(\bx^{t-1})|$ and 
\[
    \|\nabla J_{\gamma}(\xt)\| 
    \leq \|\bgbar_{\gamma}(\bxt)\| 
    + \gamma L^2 \eta \|\bgbar_{\gamma}(\bx^{t-1})\|.
\]}
\begin{theorem}
\label{approximat-loss-thm-alg-one}
\approximatlossthmalgonetext
\end{theorem}
Algorithms~\ref{app:algo:loss} and ~\ref{sec:app:algo:grad} use approximations of the objective functions 
$L_{\lambda}(\xt)$ and $J_{\gamma}(\xt)$, together with their gradients 
$\nabla L_{\lambda}(\xt)$ and $\nabla J_{\gamma}(\xt)$, 
denoted by $\abar_{\lambda}(\bxt)$, $\bgbar_{\lambda}(\xt)$, and 
$\abar_{\gamma}(\bxt), \bgbar_{\gamma}(\xt)$, respectively.  
\Cref{approximat-loss-thm-alg-one} establishes connections between the approximate and true functions, 
ensuring that the surrogate functions provide reliable estimates of the original objectives.   
Specifically, if the sequence $\bxt$ generated by Algorithm~\ref{app:algo:loss} satisfies 
$\lim_{t \to \infty} \bxt = \bx^\dagger$ and 
$\lim_{t \to \infty} \|\bgbar_{\lambda}(\xt)\| = 0$, 
then $\bx^\dagger$ is a stationary point of $L_{\lambda}$.  
Analogously, if the sequence $\bxt$ generated by Algorithm~\ref{sec:app:algo:grad} satisfies 
$\lim_{t \to \infty} \bxt = \hat{\bx}$ and 
$\lim_{t \to \infty} \|\bgbar_{\gamma}(\xt)\| = 0$, 
then $\hat{\bx}$ is a stationary point of $J_{\gamma}$.  
We present the approximate variants, FairLoss and FairGrad, here. The exact variants, FairLoss$^*$ and FairGrad$^*$, along with q-FFL, are provided in \Cref{app:exact:algorithms}.
The proof of \Cref{approximat-loss-thm-alg-one} is given in \Cref{app:Theoretical:Analysis:of:Algorithms}.

\begin{figure}[t]
\centering
\noindent
\begin{minipage}[t]{0.49\textwidth}
\refstepcounter{figure}
\label{app:algo:loss}
\vspace{0pt}
\hrule
\vspace{0.3em}
\textbf{Algorithm \thefigure: FairLoss}
\vspace{0.3em}
\hrule
\vspace{0.5em}
\begin{algorithmic}
\vspace{-0.5em}
\STATE\textbf{Input:} $\bx^0, \lambda, \eta, T$ 
\STATE\textbf{Initialize:} $a^0=0$, $\bg^0=0$
\FOR {round $t = 0,1,\ldots,T-1$}
\STATE --- \textbf{Client}-level local training -----------
\vspace{-1.0em}
\FOR {client $i = 1,\ldots,n$}
\STATE Receives $\bx^{t}, a^{t}$, and $\bg^{t}$ from the
\STATE \quad server
\STATE $a_i^t = f_i(\bxt) + \frac{\lambda}{2} \big(f_i(\bxt) - a^t \big)^2$
\STATE $\bg_i^t = \nabla f_i(\bxt)$
\STATE\hspace{2.4em} $+ \lambda \big(f_i(\bxt) - a^t\big) \big(\nabla f_i(\bxt) - \bg^t\big)$
\vspace{-1.0em}
\STATE $\bx_{i}^{t+1} = \bxt - \eta \cdot \bg_i^{t}$
\STATE Sends $\bx_{i}^{t+1}, f_i(\bxt)$, and $\nabla f_i(\bxt)$ to
\STATE \quad the server
\ENDFOR
\STATE--- \textbf{Server}-level aggregation ------------
\vspace{-1.0em}
\STATE Receives all $\bx_{i}^{t+1}, f_i(\bxt)$, and $\nabla f_i(\bxt)$
\vspace{-1.0em}
\STATE $\bx^{t+1} = \avgn \bx_{i}^{t+1}$
\STATE $a^{t+1}=\avgn f_i(\bxt)$ 
\STATE $\bg^{t+1}=\avgn \nabla f_i(\bxt)$
\STATE Sends $\bx^{t+1}, a^{t+1}$, and $\bg^{t+1}$ to the
\STATE \quad clients
\ENDFOR
\end{algorithmic}
\vspace{0.3em}
\hrule
\end{minipage}
\hfill
\begin{minipage}[t]{0.49\textwidth}
\refstepcounter{figure}
\label{sec:app:algo:grad}
\vspace{0pt}
\hrule
\vspace{0.3em}
\textbf{Algorithm \thefigure: FairGrad}
\vspace{0.3em}
\hrule
\vspace{0.5em}
\begin{algorithmic}
\vspace{-0.5em}
\STATE\textbf{Input:} $\bx^0, \gamma, \eta, T$
\STATE\textbf{Initialize:} $\bg^0=0$
\FOR {round $t = 0,1,\ldots,T-1$}
\STATE --- \textbf{Client}-level local training -----------
\vspace{-1.0em}
\FOR {client $i = 1,\ldots,n$}
\STATE Receives $\bx^{t}$ and $\bg^{t}$ from the server
\vspace{-1.0em}
\STATE $\bg_i^t = \nabla \big( f_i(\bxt) + \frac{\gamma}{2} \big\| \nabla f_i(\bxt) - \bg^t \big\|^2 \big)$
\vspace{-1.0em}
\STATE $\bx_i^{t+1} = \bxt - \eta \cdot \bg_i^t$,
\STATE Sends $\bx_{i}^{t+1}$ and $\nabla f_i(\bxt)$ to the
\STATE \quad server
\ENDFOR
\STATE--- \textbf{Server}-level aggregation ------------
\vspace{-1.0em}
\STATE $\bx^{t+1} = \avgn \bx_{i}^{t+1}$
\STATE $\bg^{t+1}=\avgn \nabla f_i(\bxt)$
\STATE Sends $\bx^{t+1}$ and $\bg^{t+1}$ to the clients
\ENDFOR
\end{algorithmic}
\vspace{0.3em}
\hrule
\end{minipage}
\end{figure}

\section{Experiments}
\label{sec:Experiments}
We evaluated the investigated methods on four widely used  image datasets:
MNIST~\cite{lecun1998gradient}, CIFAR-10, CIFAR-100~\cite{krizhevsky2009cifar}, and Tiny ImageNet~\cite{le2015tiny}, with $50$, $50$, $100$, and $100$ clients, respectively.   
To simulate varying levels of data heterogeneity among clients, we used the data generation procedure from the FL-bench framework~\cite{tam2023flbench} where a Dirichlet distribution is applied with concentration parameters $\alpha \in \{0.05, 0.1, 0.5\}$\footnote{\url{https://github.com/KarhouTam/FL-bench}}.
Smaller $\alpha$ values result in greater variability in client data distributions and more unequal dataset sizes, thereby increasing the degree of heterogeneity.
We compared the performance of the investigated methods FairLoss, FairGrad, FairLoss$^*$, FairGrad$^*$, and q-FFL against the baseline methods FedAvg, AFL, and AAggFF.
For each hyperparameter configuration, we trained all methods and selected the epoch that maximized the following validation-based criterion:
        $\overline{\mathrm{acc}} \;-\; t \cdot \sqrt{\frac{\overline{\mathrm{var}}}{m}}$,
where the criterion is computed using validation accuracies across clients with $t = 1.96$.
For clarity, for each random seed we first computed the average validation accuracy across clients and the variance of validation accuracies across clients. Then, $\overline{\text{acc}}$ denotes the average of these per-seed average accuracies, $\overline{\text{var}}$ denotes the average of these per-seed variances, and $m$ is the number of independent runs (\emph{i.e.}, seeds).
Only after selecting this best epoch and its corresponding parameters, the resulting best model was evaluated on the test data, and those test scores are the results that are reported in Tables~\ref{tab:Dirichlet-distribution:cifar100}, \ref{tab:Dirichlet-distribution:Tiny ImageNet}, \ref{tab:Dirichlet-distribution:cifar10}, and \ref{tab:Dirichlet-distribution:mnist}. 
Detailed descriptions of the datasets, partitioning strategies, model architectures, and training protocols, and hyperparameter selection are provided in \Cref{appendix:experiment:setup}.
\subsection{Results and Discussion}\label{results}
For each run, the best epoch and hyperparameters were selected based on the validation set. 
All reported results are evaluated on the test set and include test accuracy and the variance of test accuracies across clients (the fairness metric). Values in parentheses denote the standard error.
Detailed results are presented in Tables~\ref{tab:Dirichlet-distribution:cifar100}, \ref{tab:Dirichlet-distribution:Tiny ImageNet}, \ref{tab:Dirichlet-distribution:cifar10} and \ref{tab:Dirichlet-distribution:mnist}.
In these tables, the reported results correspond to different values of the Dirichlet concentration parameter $\alpha$.
For CIFAR-100 (Table~\ref{tab:Dirichlet-distribution:cifar100}), FairGrad$^*$ obtained both the best accuracy and lowest variances at $\alpha\in\{0.05,0.1\}$, and the lowest variance at $\alpha=0.5$, where FairGrad reached the highest accuracy. The regularized methods generally improved over the baseline in either accuracy, variance, or both.
For Tiny ImageNet (Table~\ref{tab:Dirichlet-distribution:Tiny ImageNet}), FairGrad$^*$ achieved the lowest variance at $\alpha\in\{0.05,0.5\}$ and the best accuracy at $\alpha=0.5$, while FairGrad and FairGrad$^*$ attained the highest accuracy at $\alpha=0.05$ and $\alpha=0.1$, respectively. Similar to the results on CIFAR-100, the regularized methods improved fairness or performance compared to the baseline. 
For CIFAR-10 (Table~\ref{tab:Dirichlet-distribution:cifar10}), FairGrad consistently outperformed the other methods, achieving the lowest variances (indicating higher fairness)  across all $\alpha$, and the highest accuracies at $\alpha\in\{0.1,0.5\}$.
Table~\ref{tab:Dirichlet-distribution:mnist} shows that for MNIST with $50$ clients, all methods perform similarly, with regularized variants offering slightly better fairness; in particular, FairGrad$^*$ achieved the lowest variance across all $\alpha$ values.

\begin{table}[H]

\caption{Performance on CIFAR-100 across Dirichlet values ($100 $ clients)}
\label{tab:Dirichlet-distribution:cifar100}
\centering
\scriptsize
\setlength{\tabcolsep}{4pt}
\begin{adjustbox}{width=\textwidth}
\renewcommand{\arraystretch}{1.2}
\begin{tabular}{lcccccc}
    \toprule
    \multirow{2}{*}{Method} 
    & \multicolumn{2}{c}{$\mathbf{Dir} (0.05)$} 
    & \multicolumn{2}{c}{$\mathbf{Dir} (0.1)$} 
    & \multicolumn{2}{c}{$\mathbf{Dir} (0.5)$} \\
    \cmidrule(lr){2-3} \cmidrule(lr){4-5} \cmidrule(lr){6-7}
    & Test Accuracy (\%) & Test Variance 
    & Test Accuracy (\%) & Test Variance 
    & Test Accuracy (\%) & Test Variance \\
    \midrule
    \text{FedAvg}      
    & 66.98~($\pm$0.73) & 0.88~($\pm$0.07) 
    & 66.69~($\pm$0.21) & 0.96~($\pm$0.04) 
    & 66.35~($\pm$0.71) & 0.92~($\pm$0.07) \\

    \text{AAggFF}        
    & 66.83~($\pm$0.50) & 0.93~($\pm$0.09) 
    & 66.68~($\pm$0.46) & 0.85~($\pm$0.08) 
    & 67.22~($\pm$0.48) & 0.89~($\pm$0.06) \\

    \text{AFL}        
    & 57.50~($\pm$0.36) & 4.94~($\pm$0.14) 
    & 61.44~($\pm$0.81) & 4.01~($\pm$0.11) 
    & 67.46~($\pm$0.46) & 2.49~($\pm$0.12) \\
    
    \text{q-FFL}        
    & 66.36~($\pm$0.32) & 0.90~($\pm$0.07) 
    & 67.70~($\pm$0.72) & 0.86~($\pm$0.02) 
    & 67.20~($\pm$0.39) & 0.92~($\pm$0.04) \\

    \text{FairLoss}    
    & 66.46~($\pm$0.78) & 0.81~($\pm$0.06) 
    & 67.55~($\pm$0.45) & 0.88~($\pm$0.06) 
    & 66.94~($\pm$0.29) & 0.81~($\pm$0.06) \\
    
    \text{FairLoss$^*$}
    & 67.16~($\pm$0.50) & 0.88~($\pm$0.05) 
    & 67.41~($\pm$0.43) & 0.90~($\pm$0.06) 
    & 67.26~($\pm$0.33) & 0.82~($\pm$0.06) \\
    
    \textbf{FairGrad}    
    & 68.90~($\pm$0.21) & 0.82~($\pm$0.03) %?
    & 69.38~($\pm$0.46) & 0.91~($\pm$0.02) 
    & \textbf{70.52}~($\pm$0.78) & 0.86~($\pm$0.04) \\
    
    \textbf{FairGrad$^*$}
    & \textbf{69.23}~($\pm$0.29) & \textbf{0.79}~($\pm$0.04)
    & \textbf{70.78}~($\pm$0.40) & \textbf{0.77}~($\pm$0.02) 
    & 69.67~($\pm$0.16) & \textbf{0.78}~($\pm$0.03) \\
    \bottomrule
\end{tabular}
\end{adjustbox}
\end{table}
\vspace{-2em}
\begin{table}[H]
\caption{Performance on Tiny ImageNet across Dirichlet values ($100 $ clients)}
\label{tab:Dirichlet-distribution:Tiny ImageNet}
\centering
\scriptsize
\setlength{\tabcolsep}{4pt}
\renewcommand{\arraystretch}{1.2}

\begin{adjustbox}{width=\textwidth}
\renewcommand{\arraystretch}{1.2}
\begin{tabular}{lcccccc}
    \toprule
    \multirow{2}{*}{Method} 
    & \multicolumn{2}{c}{$\mathbf{Dir} (0.05)$} 
    & \multicolumn{2}{c}{$\mathbf{Dir} (0.1)$} 
    & \multicolumn{2}{c}{$\mathbf{Dir} (0.5)$} \\
    \cmidrule(lr){2-3} \cmidrule(lr){4-5} \cmidrule(lr){6-7}
    & Test Accuracy (\%) & Test Variance 
    & Test Accuracy (\%) & Test Variance 
    & Test Accuracy (\%) & Test Variance \\
    \midrule
    \text{FedAvg}      
    & 59.03~($\pm$0.10) & \textbf{0.47}~($\pm$0.01) 
    & 59.39~($\pm$0.27) &\textbf{ 0.53}~($\pm$0.01) 
    & 59.42~($\pm$0.15) & 0.44~($\pm$0.02) \\

    \text{AAggFF}        
    & 59.15~($\pm$0.23) & 0.51~($\pm$0.02) 
    & 59.30~($\pm$0.16) & 0.57~($\pm$0.02) 
    & 59.70~($\pm$0.25) &0.49~($\pm$0.02) \\

    \text{AFL}        
    & 59.27~($\pm$1.30) & 3.67~($\pm$0.17) 
    & 58.88~($\pm$1.55) & 2.51~($\pm$0.19) 
    & 55.16~($\pm$0.52) &1.69~($\pm$0.07) \\
    
    \text{q-FFL}        
    & 58.77~($\pm$0.12) & 0.56~($\pm$0.02) 
    & 59.25~($\pm$0.12) & 0.56~($\pm$0.02) 
    & 59.71~($\pm$0.01) & 0.47~($\pm$0.01) \\
    
    \text{FairLoss}    
    & 59.10~($\pm$0.15) & 0.49~($\pm$0.03) 
    & 59.60~($\pm$0.39) & \textbf{0.53}~($\pm$0.01) 
    & 59.50~($\pm$0.03) & 0.46~($\pm$0.03) \\
    
    \text{FairLoss$^*$}
    & 59.15~($\pm$0.31) & 0.49~($\pm$0.03) 
    & 59.55~($\pm$0.17) & 0.54~($\pm$0.04) 
    & 59.62~($\pm$0.15) & 0.46~($\pm$0.01) \\
    
    \textbf{FairGrad}    
    & \textbf{62.06}~($\pm$0.11) & 0.52~($\pm$0.01) 
    & 62.15~($\pm$0.85) & 0.56~($\pm$0.02) 
    & 60.01~($\pm$0.02) & 0.44~($\pm$0.02) \\
    
    \textbf{FairGrad$^*$}
    & 59.92~($\pm$0.40) & \textbf{0.47}~($\pm$0.02)
    & \textbf{63.46}~($\pm$0.12) & 0.56~($\pm$0.01) 
    & \textbf{62.84}~($\pm$0.11) &  \textbf{0.36}~($\pm$0.01) \\
    \bottomrule
\end{tabular}
\end{adjustbox}
\end{table}
\vspace{-2em}
\begin{table}[H]
\caption{Performance on CIFAR-10 across Dirichlet values ($50$ clients)}
\label{tab:Dirichlet-distribution:cifar10}
\centering
\scriptsize
\setlength{\tabcolsep}{4pt}
\renewcommand{\arraystretch}{1.2}
\begin{adjustbox}{width=\textwidth}
\renewcommand{\arraystretch}{1.2}
\begin{tabular}{lcccccc}
    \toprule
    \multirow{2}{*}{Method} 
    & \multicolumn{2}{c}{$\mathbf{Dir} (0.05)$} 
    & \multicolumn{2}{c}{$\mathbf{Dir} (0.1)$} 
    & \multicolumn{2}{c}{$\mathbf{Dir} (0.5)$} \\
    \cmidrule(lr){2-3} \cmidrule(lr){4-5} \cmidrule(lr){6-7}
    & Test Accuracy (\%) & Test Variance 
    & Test Accuracy (\%) & Test Variance 
    & Test Accuracy (\%) & Test Variance \\
    \midrule
    \text{FedAvg}      
    & 90.29~($\pm$0.26) & 1.16~($\pm$0.03) 
    & 84.89~($\pm$0.36) & 1.50~($\pm$0.14) 
    & 69.44~($\pm$0.86) & 0.75~($\pm$0.06) \\

    \text{AAggFF}      
    & 90.53~($\pm$0.15) & 1.01~($\pm$0.03) 
    & 84.81~($\pm$0.36) & 1.43~($\pm$0.06) 
    & 68.45~($\pm$0.64) & 0.80~($\pm$0.03) \\

     \text{AFL}      
    & \textbf{93.49}~($\pm$0.27) & 1.30~($\pm$0.01)
    & 72.77~($\pm$1.58) & 2.07~($\pm$0.43) 
    & 63.75~($\pm$1.34) & 0.79~($\pm$0.23) \\
    
    \text{q-FFL}        
    & 90.05~($\pm$0.22) & 1.14~($\pm$0.03) 
    & 84.58~($\pm$0.33) & 1.52~($\pm$0.09) 
    & 70.35~($\pm$0.42) & 0.65~($\pm$0.05) \\
    
    \text{FairLoss}    
    & 90.40~($\pm$0.11) & 1.14~($\pm$0.05) 
    & 84.71~($\pm$0.31) & 1.51~($\pm$0.09) 
    & 69.70~($\pm$0.51) & 0.74~($\pm$0.08) \\
    
    \text{FairLoss$^*$}
    & 90.21~($\pm$0.22) & 1.10~($\pm$0.06) 
    & 85.31~($\pm$0.12) & 1.45~($\pm$0.08) 
    & 69.84~($\pm$0.36) & 0.77~($\pm$0.03) \\
    
    \textbf{FairGrad}    
    & {92.18}~($\pm$0.10) & \textbf{0.73}~($\pm$0.02) 
    & \textbf{86.70}~($\pm$0.11) & \textbf{1.11}~($\pm$0.09) 
    & \textbf{72.61}~($\pm$0.44) & \textbf{0.62}~($\pm$0.01) \\
    
    \textbf{FairGrad$^*$}
    & 90.58~($\pm$0.14) & 1.12~($\pm$0.02)
    & 84.82~($\pm$0.16) & 1.55~($\pm$0.09) 
    & 69.46~($\pm$0.41) & 0.75~($\pm$0.04) \\
    \bottomrule
\end{tabular}
\end{adjustbox}
\end{table}
\vspace{-2em}
\begin{table}[H]
\caption{Performance on MNIST across Dirichlet values ($50$ clients)}
\label{tab:Dirichlet-distribution:mnist}
\centering
\scriptsize
\setlength{\tabcolsep}{4pt}
\renewcommand{\arraystretch}{1.2}

\begin{adjustbox}{width=\textwidth}
\renewcommand{\arraystretch}{1.2}
\begin{tabular}{lcccccc}
    \toprule
    \multirow{2}{*}{Method} 
    & \multicolumn{2}{c}{$\mathbf{Dir} (0.05)$}
    & \multicolumn{2}{c}{$\mathbf{Dir} (0.1)$}
    & \multicolumn{2}{c}{$\mathbf{Dir} (0.5)$} \\
    \cmidrule(lr){2-3} \cmidrule(lr){4-5} \cmidrule(lr){6-7}
    & Test Accuracy (\%) & Test Variance 
    & Test Accuracy (\%) & Test Variance 
    & Test Accuracy (\%) & Test Variance \\
    \midrule
    \text{FedAvg}      
    & 94.42~($\pm$0.18) & 0.07~($\pm$0.07) 
    & 94.10~($\pm$0.20) & 0.08~($\pm$0.08) 
    & 88.99~($\pm$0.18) & \textbf{0.05}~($\pm$0.05) \\

    \text{AAggFF}      
    & \textbf{95.12}~($\pm$0.15) & 0.07~($\pm$0.06) 
    & \textbf{94.22}~($\pm$0.30) & 0.09~($\pm$0.06) 
    & 89.04~($\pm$0.20) & \textbf{0.05}~($\pm$0.04) \\
    
    \text{AFL}      
    & 94.54~($\pm$0.04) & 0.30~($\pm$0.07) 
    & 93.57~($\pm$0.02) & 0.28~($\pm$0.01) 
    & 88.86~($\pm$0.03) & {0.08}~($\pm$0.03) \\
    
    \text{q-FFL}        
    & 95.01~($\pm$0.20) & 0.06~($\pm$0.07) 
    & 94.10~($\pm$0.22) & 0.08~($\pm$0.07) 
    & 89.07~($\pm$0.19) & \textbf{0.05}~($\pm$0.05) \\
    
    \text{FairLoss}    
    & 94.42~($\pm$0.22) & 0.07~($\pm$0.06) 
    & 94.10~($\pm$0.19) & 0.08~($\pm$0.08) 
    & 88.00~($\pm$0.18) & \textbf{0.05}~($\pm$0.05) \\
    
    \text{FairLoss$^*$}
    & 94.31~($\pm$0.22) & \textbf{0.05}~($\pm$0.06) 
    & 94.12~($\pm$0.19) & 0.08~($\pm$0.09) 
    & 88.99~($\pm$0.18) & \textbf{0.05}~($\pm$0.05) \\
    
    \textbf{FairGrad}    
    & 94.31~($\pm$0.19) & \textbf{0.05}~($\pm$0.06) 
    & 94.09~($\pm$0.20) & 0.08~($\pm$0.08) 
    & \textbf{89.10}~($\pm$0.14) & \textbf{0.05}~($\pm$0.04) \\
    
    \textbf{FairGrad$^*$}
    & 94.31~($\pm$0.19) & \textbf{0.05}~($\pm$0.05)
    & 94.08~($\pm$0.25) & \textbf{0.07}~($\pm$0.07) 
    & 89.00~($\pm$0.18) & \textbf{0.05}~($\pm$0.08) \\
    \bottomrule
\end{tabular}
\end{adjustbox}
\end{table}

\section{Conclusion}
\label{sec:Conclusion}
In this work, we focused on fairness-aware FL algorithms that explicitly reduce client disparities through variance-based regularization of client losses or gradients.
We evaluated several methods in this category, including a gradient variance regularization approach, proposed here for the first time in the context of performance equitable fairness, and implemented in two variants:
An approximate version that uses the average gradient computed at the previous iteration, and an exact version that uses the average gradient from the current iteration across clients. The exact variant can be useful when the task is more challenging or client gradients vary substantially, whereas the approximate version is preferable when communication efficiency is the main concern.

The theoretical analysis indicates that, under the sufficient conditions identified in our analysis, the regularized objectives have different stationary points compared to the unregularized ones. Empirical results confirm that, in heterogeneous settings with more clients and complex datasets, the regularized methods improve both fairness and overall model performance, with the gradient alignment variants consistently delivering strong results.
In contrast, in homogeneous settings, the theoretical results suggest minimal to no differences between these methods, which aligns with the empirical observations. 
The experiments focused on image classification benchmarks with full client participation; to evaluate other tasks and partial client participation is an important direction for future work.
Overall, this work improves our understanding of variance-based fairness regularization in general and in particular highlights the proposed gradient alignment regularizer as a principled and effective approach to improve fairness in FL.

\begin{credits}
\subsubsection{\ackname} 
This work was supported by the Swedish Cancer Society (grant number 22 2428 Pj) and Lion's Cancer Research Foundation in Northern Sweden (grant numbers LP 22-2319, LP 24-2367, and AMP 26-1265). The computations were enabled by resources provided by the National Academic Infrastructure for Supercomputing in Sweden (NAISS) at C3SE and HPC2N, partially funded by the Swedish Research Council through grant agreement no.~2022-06725. Ali Dadras was supported by the Wallenberg AI, Autonomous Systems and Software Program (WASP), funded by the Knut and Alice Wallenberg Foundation, and by the Swedish Research Council under grant number 2023-05476.
We are also grateful for valuable discussions with Alp Yurtsever.
\subsubsection{\discintname}
The authors have no competing interests to declare that are relevant to the content of this article.
\end{credits}
% ----------------------------------------------------------
\bibliographystyle{splncs04}

\bibliography{bibliography}

\clearpage
\appendix

\begin{center}
{\Large \textbf{Appendix}}\\[0.5em]
{\Large \textbf{FAIRVAR: Fair Federated Learning via Variance Regularization}}
\end{center}
\label{appendix}

\section{Analysis of Data Distribution}\label{appendix}
We begin this section by stating the key assumptions and definitions that form the foundation of the theoretical analysis.
Specifically, we assume that the local loss functions, $f_i(\bx)$, are smooth, and  the number of clients satisfies $n \geq 1$. The following definition formalizes how we characterize the stationary points and minimizers of the objective functions considered in our study.

\begin{definition}[minimizers]
    \label{app:def:minimizers}
    For the functions $F(\bx)$, $L_\lambda(\bx)$, $J_{\gamma}(\bx)$, and $H_q(\bx)$,  we define 
     \begin{align*}
    &\bxstar \in \argmin_{\bx \in \R^d} ~ F(\bx),   \  \ \text{such that} \ \nabla{F}(\bxstar) = 0
    \\
    &\bxlambdastar \in \argmin_{\bx \in \R^d} ~ L_{\lambda} (\bx), \  \ \text{such that} \ \nabla{L}_{\lambda}(\bxlambdastar) = 0
    \\
    &\bxgstar \in \argmin_{\bx \in \R^d} ~ J_{\gamma}(\bx), \  \ \text{such that} \ \nabla{J}_{\gamma}(\bxgstar) = 0
    \\
    &\bxqstar \in \argmin_{\bx \in \R^d} ~   H_q(\bx), \  \ \text{such that} \ \nabla{H}_q(\bxqstar) = 0.
    \end{align*}
\end{definition}

\subsection{Homogeneous Case}\label{homogenous:data}

We now prove Theorem~\ref{qffl-relation-loss}. We recall the theorem:

\medskip
\noindent\textbf{Theorem~\ref{qffl-relation-loss}.}
\textit{\qfflrelationlosstext}

\begin{proof}
From \Cref{qffl}, for $q=1, r_i=\frac{1}{n}$, we have 
\begin{equation*}
    \E\big[H_1(\bx)\big]
        = \E\left[\avgcnn{1} \left( f_i(\bx) \right)^{2}\right].
\end{equation*}
Using the standard variance identity,
\begin{equation*}
    \text{Var}\big(f_i(\bx)\big)
        = \E\big[f_i(\bx)^2\big] - \E\big[f_i(\bx)\big]^2,
\end{equation*}
we have
\begin{equation}\label{eq:H1}
    \E\big[H_1(\bx)\big]
        = \E\left[\avgcnn{1} f_i(\bx)^2\right]
        = \avgcnn{1} \E\big[f_i(\bx)^2\big]
        = \avgcnn{1} \text{Var}\big(f_i(\bx)\big) + \E\big[f_i(\bx)\big]^2.
\end{equation}
Now using the sample variance of client losses, we have 
\begin{equation}\label{eq:H1:H1}
\begin{aligned}
\avgcnn{1} \,\mathrm{Var}\big(f_i(\bx)\big)
&= \frac{n}{2n}\frac{1}{n-1}\sum_{i=1}^{n}
\mathbb{E}\Big[\big(f_i(\bx) - F(\bx)\big)^2\Big] \\
&= \frac{n}{n-1}
\bigg(\frac{1}{2n}\sum_{i=1}^{n}
\mathbb{E}\Big[\big(f_i(\bx) - F(\bx)\big)^2\Big]\bigg).
\end{aligned}
\end{equation}
By assumption, $\E[f_i(\bx)] \leq C, \forall i$, so $\E[f_i(\bx)]^2  \leq C\E[f_i(\bx)]$. Using this fact and \Cref{eq:H1:H1}, we can write \Cref{eq:H1} as
\begin{align}\label{eq:H1:H1:H1}
    \E\big[H_1(\bx)\big]
        &= \avgcnn{1} \E\big[ f_i(\bx) \big]^2 + \frac{n}{n-1}\bigg(\frac{1}{2n}\sum_{i=1}^{n}\E\Big[\big(f_i(\bx) - F(\bx)\big)^2\Big]\bigg) \\
        &\leq \avgcn{1} C\E\big[ f_i(\bx) \big] + \frac{n}{n-1} \bigg(\frac{1}{2n}\sum_{i=1}^{n}\E\Big[\big(f_i(\bx) - F(\bx)\big)^2\Big]\bigg) \\
        &= C\E\big[F(\x)\big] + \frac{C n}{C (n-1)} \bigg(\frac{1}{2n}\sum_{i=1}^{n}\E\Big[\big(f_i(\bx) - F(\bx)\big)^2\Big]\bigg) \\ 
        &= C\E\big[F(\x)\big] + \frac{C \lambda}{2n}\sum_{i=1}^{n}\E\Big[\big(f_i(\bx) - F(\bx)\big)^2\Big] \\ 
        &= C\E\big[L_{\lambda}(\bx)\big],
\end{align}
with $\lambda=\frac{n}{C(n-1)}$.
Hence, we see that $\E [H_q(\bx)] \leq C\E[L_{\lambda}(\bx)]$,
when $q=1$ and $\lambda=\frac{n}{C(n-1)}$.
\end{proof}
% ---------------------------------------------------------

\subsection{Heterogeneous Case}\label{heterogeneous:data}

We now prove Theorem~\ref{stationary-point-thm-three}. We recall the theorem:

\medskip
\noindent\textbf{Theorem~\ref{stationary-point-thm-three}.}
\textit{\stationarypointthmthreetext}

\begin{proof}
Given the assumption that $\bxstar \in \argmin_{\bx \in \R^d} ~ F(\bx)$, and 
$\nabla F(\bxstar)= 0$, we want to show that $\nabla L_\lambda(\bxstar)=\nabla J_{\gamma}(\bxstar) =\nabla H_q(\bxstar) =0$. Using the gradient of \Cref{eq:fairFL}, 
we obtain
\begin{equation}\label{stationary-point-eq-nine}
    \nabla L_\lambda(\bxstar)
    =\underbrace{\nabla F(\bxstar)}_{=0}
    +\avgcn{\lambda} \big(f_i(\bxstar)-F(\bxstar)\big)\big(\nabla f_i(\bxstar)-\nabla F(\bxstar)\big).
\end{equation}
It follows that if $f_i(\bxstar)=f_j(\bxstar)$ or $\nabla f_i(\bxstar)=\nabla f_j(\bxstar)$ for all $i,j$, then $F(\bxstar)=f_i(\bxstar)$ or $\gradF(\bxstar)=\gradfi(\bxstar)$ for all $i$, resulting in $\nabla L_\lambda(\bxstar)=0$.
Similarly, using the gradient of \Cref{eq:fairGrad}, we have
\begin{equation}
    \nabla J_\gamma(\bxstar)
    =\underbrace{\nabla F(\bxstar)}_{=0}
    + \avgcn{\lambda} \underbrace{\big\|\nabla f_i(\bxstar)-\nabla F(\bxstar)\big\|}_{=0}\norm{\nabla^2f_i(\bxstar)-\nabla^2F(\bxstar)}=0.
\end{equation}
Now we show that $\nabla H_q(\bxstar) = 0$. From \Cref{qffl},
\begin{equation}
    \nabla H_q(\bxstar) = \sum_{i=1}^{n} r_i f_i^q(\bxstar)\nabla f_i(\bxstar).
\end{equation}
If all losses or gradients are equal, then form 
$\gradF(\bxstar)=\avgcn{1}\gradfi(\bxstar)=\gradfi(\bxstar)=0$,  we have $\nabla H_q(\bxstar)=0$.
Now we show that with the similar assumptions if $f_i(\bxlambdastar)=f_j(\bxlambdastar)$ or $\nabla f_i(\bxlambdastar)=\nabla f_j(\bxlambdastar)$ for all $i,j$, then $\nabla F(\bxlambdastar)=0$.  Since $\bxlambdastar \in \argmin_{\bx \in \R^d} ~ L_\lambda(\bx)$, and $\nabla  L_\lambda(\bxlambdastar)=0$, we have
\begin{equation}\label{stationary-point-eq-ten}
0=\nabla L_\lambda(\bxlambdastar)
=\nabla F(\bxlambdastar)
+ \avgcn{\lambda}\big(f_i(\bxlambdastar) - F(\bxlambdastar)\big)\big(\nabla f_i(\bxlambdastar) - \nabla F(\bxlambdastar)\big).
\end{equation}
Similar discussion as above shows that if $f_i(\bxlambdastar)=f_j(\bxlambdastar)$ or $\nabla f_i(\bxlambdastar)=\nabla f_j(\bxlambdastar)$ for all $i,j$, then 
\begin{equation*}
\avgcn{\lambda}\big(f_i(\bxlambdastar) - F(\bxlambdastar)\big)\big(\nabla f_i(\bxlambdastar) - \nabla F(\bxlambdastar)\big)=0.
\end{equation*}
 Thus, $\nabla L_\lambda(\bxlambdastar)
=\nabla F(\bxlambdastar)=0$. Same discussion holds for $\bxgstar $, and $\bxqstar$, and we have $\nabla F(\bxgstar)=0$ and $\nabla F(\bxqstar)=0$.
\end{proof}
We now prove Theorem~\ref{stationary-point-thm-one}. We recall the theorem:

\medskip
\noindent\textbf{Theorem~\ref{stationary-point-thm-one}.}
\textit{\stationarypointthmonetext}

\begin{proof}
According to the assumption, for some $1\leq i, j\leq n$,  $f_i(\bxstar) \neq f_j(\bxstar)$ and $\nabla f_i(\bxstar) \neq 0$. 
Since $\nabla F(\bxstar)=0$, then $\sum _{i=1}^n\nabla f_i(\bxstar)=0$ and this means that $\nabla f_1(\bxstar), \ldots, \nabla f_n(\bxstar)$ are linearly dependent vectors.  We will begin by assuming that $m=n-1$. Without loss of generality, assume that $\nabla f_1(\bxstar), \ldots, \nabla f_{n-1}(\bxstar)$ are linearly independent and $\nabla f_1(\bxstar), \ldots, \nabla f_{n-1}(\bxstar), \nabla f_{n}(\bxstar)$ are linearly dependent. From \Cref{eq:fairFL}, we have
\begin{align}\label{stationary-point-eq-one}
    \nabla L_\lambda(\bxstar)
    &=\underbrace{\nabla F(\bxstar)}_{=0}
    +\avgcn{\lambda} \big(f_i(\bxstar)-F(\bxstar)\big)\big(\nabla f_i(\bxstar)-\underbrace{\nabla F(\bxstar)}_{=0}\big)\notag \\
    &=\avgcn{\lambda}\big(f_i(\bxstar) - F(\bxstar)\big)\nabla f_i(\bxstar).
\end{align}
Substituting the $\nabla f_n(\bxstar)=-\sum_{i=1}^{n-1}\nabla f_i(\bxstar)$ in \Cref{stationary-point-eq-one}, we have that
\begin{align}\label{stationary-point-eq-two}
    \nabla L_\lambda(\bxstar)
    &=\frac{\lambda}{n}\sum_{i=1}^{n-1}\big(f_i(\bxstar) - F(\bxstar)\big)\nabla f_i(\bxstar) -  \big(f_n(\bxstar) - F(\bxstar)\big)\nabla f_i(\bxstar) \notag\\
    &=\frac{\lambda}{n}\sum_{i=1}^{n-1}\big(f_i(\bxstar) - f_n(\bxstar)\big)\nabla f_i(\bxstar).
\end{align}
By assumption for some $i$, $f_i(\bxstar) - f_n(\bxstar)\neq 0$ and $\nabla f_1(\bxstar), \ldots, \nabla f_{n-1}(\bxstar)$ are linearly independent, so $\nabla L_\lambda(\bxstar)\neq 0$. 
Let $1<m<n-1$. Assume without loss of generality that $\nabla f_1(\bxstar), \ldots, \nabla f_{m}(\bxstar)$, are linearly independent and $\nabla f_1(\bxstar), \ldots, \nabla f_{m}(\bxstar), \nabla f_{m+1}(\bxstar), \ldots,\nabla f_{n}(\bxstar)$ are linearly dependent. By assumption we have $f_{m+1}(\bxstar)=\cdots=f_{n}(\bxstar)=C$. 
Using $\sum_{i=1}^n\nabla f_i(\bxstar)=0$ and substituting $\sum_{i=m+1}^n\nabla f_i(\bxstar)=-\sum_{i=1}^{m}\nabla f_i(\bxstar)$ into \Cref{stationary-point-eq-one}, we obtain
\begin{align}
    \nabla L_\lambda(\bxstar)
        &= \frac{\lambda}{n}\sum_{i=1}^{m}\big(f_i(\bxstar) - F(\bxstar)\big)\nabla f_i(\bxstar) -  \big(C - F(\bxstar)\big)\nabla f_i(\bxstar) \nonumber\\
        &= \frac{\lambda}{n}\sum_{i=1}^{m}\big(f_i(\bxstar) - C\big)\nabla f_i(\bxstar). \nonumber
\end{align}
By assumption for at least one index $1\leq i\leq m$, we have $f_i(\bxstar) - C\neq 0$ and $\nabla f_1(\bxstar), \ldots, \nabla f_{m}(\bxstar)$ are linearly independent, then $\nabla L_\lambda(\bxstar)\neq 0$. 

Now we consider the second case in the theorem. Without loss of generality, we assume that $\nabla f_1(\bxstar), \ldots, \nabla f_{m}(\bxstar)$, are non-zero gradient vectors, and that the loss functions satisfy $C=f_1(\bxstar)=f_2(\bxstar)=\cdots=f_{m-1}(\bxstar)\neq f_m(\bxstar)$. Therefore,
\begin{align*}
\nabla L_\lambda(\bxstar)
&=\frac{\lambda}{n}\sum_{i=1}^{m}\big(f_i(\bxstar) - F(\bxstar)\big)\nabla f_i(\bxstar) \notag \\
&=\frac{\lambda}{n}\sum_{i=1}^{m-1}\big(C - F(\bxstar)\big)\nabla f_i(\bxstar) + \frac{\lambda}{n}\big(f_{m}(\bxstar) - F(\bxstar)\big)\nabla f_m(\bxstar).
\end{align*}
Now using the fact that $\sum_{i=1}^{n}\nabla f_i(\bxstar)=\sum_{i=1}^{m}\nabla f_i(\bxstar)=0$, and  we have 
 $\sum_{i=1}^{m-1}\nabla f_i(\bxstar) = -\nabla f_m(\bxstar)$, then
\begin{align*}
\nabla L_\lambda(\bxstar)
&=\frac{\lambda}{n}\big(C - F(\bxstar)\big)\sum_{i=1}^{m-1}\nabla f_i(\bxstar) + \frac{\lambda}{n}\big(f_{m}(\bxstar) - F(\bxstar)\big)\nabla f_m(\bxstar) \notag \\
&= \frac{\lambda}{n}\underbrace{\big(f_{m}(\bxstar) - C \big)}_{\neq 0}\underbrace{\nabla f_m(\bxstar)}_{\neq 0}\neq 0.
\end{align*}
Now we prove that $ \nabla F (\bxlambdastar) \neq 0$.
Based on the assumption, for some $1\leq i, j\leq n$,  $f_i(\bxlambdastar) \neq f_j(\bxlambdastar)$ and $\nabla f_i(\bxlambdastar) \neq 0$. From \Cref{eq:fairFL}, we have
\begin{align}\label{stationary-point-eq-five}
    \nabla L_\lambda(\bxlambdastar)
    &= \nabla F(\bxlambdastar)
    + \avgcn{\lambda} \big(f_i(\bxlambdastar) - F(\bxlambdastar)\big)\big(\nabla f_i(\bxlambdastar) - \nabla F(\bxlambdastar)\big)\notag \\
    &=\avgn \underbrace{\big(1 + \lambda(f_i(\bxlambdastar) - F(\bxlambdastar))\big)}_{=:\beta_i}\nabla f_i(\bxlambdastar)=0,
\end{align}
where \Cref{stationary-point-eq-five} indicates that not all
$\beta_i$ can be zero. This is because if all  $\beta_i=0$, then $\sum_{i=1}^{n} \beta_i=0$. Hence, 
\begin{equation*}
    0=\sum_{i=1}^{n} \beta_i=n+ \lambda \underbrace{\sum_{i=1}^{n}f_i(\bxlambdastar) - F(\bxlambdastar)}_{=0}=n.
\end{equation*}
This is a contradiction with the fact that $n\neq 0$.
This implies that the gradients of the local loss functions at $\bxlambdastar$, $\nabla f_1(\bxlambdastar), \ldots, \nabla f_n(\bxlambdastar)$ are linearly dependent. Let $m=n-1$. Without loss of generality, assume that $\nabla f_1(\bxlambdastar), \ldots, \nabla f_{n-1}(\bxlambdastar)$ are linearly independent and $\nabla f_1(\bxlambdastar), \ldots, \nabla f_{n-1}(\bxlambdastar), \nabla f_{n}(\bxlambdastar)$ are linearly dependent and $\beta_n\neq 0$. From \Cref{stationary-point-eq-five}, $\sum_{i=1}^{n}\beta_i\nabla f_i(\bxlambdastar)=0$. It follows that
\begin{equation}\label{stationary-point-eq-seven}
    \beta_n\nabla f_n(\bxlambdastar) = -\sum_{i=1}^{n-1}\beta_i\nabla f_i(\bxlambdastar).
\end{equation}
Then
\begin{align}
\nabla F(\bxlambdastar)
&= \avgn \nabla f_i(\bxlambdastar) \\
&= \frac{1}{n}\sum_{i=1}^{n-1}\nabla f_i(\bxlambdastar)
   - \frac{1}{n} \sum_{i=1}^{n-1}\frac{\beta_i}{\beta_n}\nabla f_i(\bxlambdastar) \\
&= \frac{1}{n}\sum_{i=1}^{n-1} \bigg(1 - \frac{\beta_i}{\beta_n}\bigg)\nabla f_i(\bxlambdastar).
\end{align}
Since we have assumed that $\nabla f_1(\bxlambdastar), \ldots, \nabla f_{n-1}(\bxlambdastar)$ are linearly independent and at least for one $1\leq i \leq n-1$, $\beta_i\neq \beta_n$, therefore $\nabla F(\bxlambdastar)\neq 0$.
Here we show that $\beta_i \neq \beta_n$. If $\beta_i = \beta_n$ for all $1\leq i \leq n-1$, from \Cref{stationary-point-eq-five} it would imply that $f_i(\bxlambdastar) = f_n(\bxlambdastar)$ for all $1\leq i \leq n-1$. However, this is not the case we are considering. 

Let there be $1 < m < n-1$ linearly independent gradient vectors. Without loss of generality, we assume that $\nabla f_1(\bxlambdastar), \ldots, \nabla f_{m}(\bxlambdastar)$ are linearly independent and  $\nabla f_1(\bxlambdastar), \ldots,$ $\nabla f_{m}(\bxlambdastar), \nabla f_{m+1}(\bxlambdastar), \ldots, \nabla f_{n}(\bxlambdastar),$ are linearly dependent. By assumption $ f_{m+1}(\bxlambdastar) = \cdots = f_{n}(\bxlambdastar)$, so we have $\beta_{m+1}=\dots=\beta_{n}=\beta\neq 0$. From \Cref{stationary-point-eq-five},
\begin{equation}
    \sum_{i=m+1}^{n}\beta\nabla f_i(\bxlambdastar) = -\sum_{i=1}^{m}\beta_i\nabla f_i(\bxlambdastar).
\end{equation}
Then
\begin{equation}\label{stationary-point-eq-eight}
\begin{aligned}
\nabla F(\bxlambdastar)
&= \avgn \nabla f_i(\bxlambdastar) \\
&= \frac{1}{n}\sum_{i=1}^{m}\nabla f_i(\bxlambdastar)
   - \frac{1}{n} \sum_{i=1}^{m}\frac{\beta_i}{\beta}\nabla f_i(\bxlambdastar) \\
&= \frac{1}{n}\sum_{i=1}^{m}
\left(1 - \frac{\beta_i}{\beta}\right)\nabla f_i(\bxlambdastar).
\end{aligned}
\end{equation}
Since we have assumed that $\nabla f_1(\bxlambdastar), \ldots, \nabla f_{m}(\bxlambdastar)$ are linearly independent and, through a similar discussion for the case of $n-1$ linearly independent gradients, for at least one $1\leq i \leq m$, $\beta_i\neq \beta$. Therefore, $\nabla F(\bxlambdastar)\neq 0$.

 Now we consider the second case in the theorem. Without loss of generality, we assume that $\nabla f_1(\bxlambdastar), \ldots, \nabla f_{m}(\bxlambdastar)$, are non-zero gradient vectors, and that the  loss functions satisfy $f_1(\bxlambdastar)=f_2(\bxlambdastar)=\cdots=f_{m-1}(\bxlambdastar)\neq f_m(\bxlambdastar)$. Therefore we have, $0 \neq \beta_{1}=\dots=\beta_{m-1}=\beta\neq \beta_{m}$, and using \Cref{stationary-point-eq-five}, $\sum_{i=1}^{n}\beta_i\nabla f_i(\bxlambdastar)=\sum_{i=1}^{m}\beta_i\nabla f_i(\bxlambdastar)=0$.
It follows that 
\begin{equation}\label{stationary-point-eq-seven_two}
    \beta_m\nabla f_m(\bxlambdastar) = -\sum_{i=1}^{m-1}\beta_i\nabla f_i(\bxlambdastar)= -\sum_{i=1}^{m-1}\beta\nabla f_i(\bxlambdastar).
\end{equation}
Then using \Cref{stationary-point-eq-seven_two},
\begin{align}\label{stationary-point-eq-eight_two}
    \nabla F(\bxlambdastar) = \frac{1}{n} \sum_{i=1}^{n} \nabla f_i(\bxlambdastar) 
    = \frac{1}{n} \sum_{i=1}^{m} \nabla f_i(\bxlambdastar)  
    &= \frac{1}{n} \left( \sum_{i=1}^{m-1} \nabla f_i(\bxlambdastar) + \nabla f_m(\bxlambdastar) \right) \nonumber \\
    &= \frac{1}{n} \left( -\frac{\beta_m}{\beta} \nabla f_m(\bxlambdastar) + \nabla f_m(\bxlambdastar) \right) \nonumber \\
    &= \frac{1}{n} \bigg(1 - \frac{\beta_m}{\beta} \bigg) \nabla f_m(\bxlambdastar).
\end{align}
Since $\beta \neq \beta_m$ and $\nabla f_m(\bxlambdastar)\neq 0$, then $\nabla F(\bxlambdastar) \neq 0$.
\end{proof}

\noindent We now prove Theorem~\ref{stationary-point-thm-four}. We recall the theorem:

\medskip
\noindent\textbf{Theorem~\ref{stationary-point-thm-four}.}
\textit{\stationarypointthmfourtext}

\begin{proof}
From \Cref{stationary-point-thm-three}, we know that $\nabla F( \bxstar)=\nabla L_\lambda(\bxstar)=0$.
Since $\bar{o}=\avgcn{1}o_i$, so we can write $\hat{F}(\bx)=F(\bx) + \bar{o}$ and
\begin{equation}\label{stationary-point-eq-eleven}
    \hat{L}_{\lambda}(\bx)=\hat{F}(\bx)+ \avgcnn{\lambda} \big(\hat{f}_i(\bx) - \hat{F}(\bx) \big)^2=F(\bx) + \bar{o}+\avgcnn{\lambda}\big({f}_i(\bx) - {F}(\bx) + o_i-\bar{o} \big)^2.
\end{equation}
By calculating the gradient of function $\hat{L}_{\lambda}(\bxstar)$, and using the facts that $\nabla F(\bxstar)=0$ and $F(\bxstar)=f_i(\bxstar)$ for all $i$, we have
\begin{align}\label{stationary-point-eq-twelve}
    \nabla \hat{L}_{\lambda}(\bxstar)
    &= \underbrace{\nabla F(\bxstar)}_{=0}
    + \avgcn{\lambda}\big(\underbrace{f_i(\bxstar) - F(\bxstar)}_{=0} + o_i-\bar{o}\big)\big(\nabla f_i(\bxstar) - \underbrace{\nabla F(\bxstar)}_{=0}\big)\notag \\
    &=\avgcn{\lambda}\big(o_i-\bar{o}\big)\nabla f_i(\bxstar).
\end{align}
Without loss of generality, we assume that $\nabla f_1(\bxstar), \ldots ,\nabla f_m(\bxstar)$ are non zero gradients. Since $\nabla F(\bxstar)=0$, we can write $\nabla f_m(\bxstar)= -\sum_{i=1}^{m-1} \nabla f_i(\bxstar)$. Assume that $o_i=o_j=o$ for $i,j\in\{1,\ldots,m-1\}$.
Then, from \Cref{stationary-point-eq-twelve} and substituting the $\sum_{i=1}^{m-1}\nabla f_i(\bxstar)=-\nabla f_m(\bxstar)$, we have
\begin{align*}
    \nabla \hat{L}_{\lambda}(\bxstar)=\avgcn{\lambda} o_i\nabla f_i(\bxstar)
         &= \frac{\lambda}{n}\sum_{i=1}^{m} o_i\nabla f_i(\bxstar)\\
        &= \frac{\lambda}{n}\bigg(\sum_{i=1}^{m-1} o_i\nabla f_i(\bxstar) + o_m\nabla f_m(\bxstar)\bigg) \notag \\
        &= \frac{\lambda}{n}(-o + o_m)\nabla f_m(\bxstar).
\end{align*}
Now if we choose $o\neq o_m$, then $\nabla  \hat{L}_{\lambda}(\bxstar)\neq 0$, and  $\bxstar$ is not a stationary point of the function $\hat{L}_{\lambda}$.

\end{proof}

\newcommand{\stationarypointthmfivetext}{

   Assume that the $f_i$'s are differentiable. Let 
$\bxstar \in \argmin_{\bx \in \mathbb{R}^d} F(\bx)$ and 
$\bxlambdastar \in \argmin_{\bx \in \mathbb{R}^d} L_{\lambda}(\bx)$. 
Then,
\[
\avgcn{1} \big(f_i(\bxlambdastar) - F(\bxlambdastar)\big)^2 
\leq 
\avgcn{1} \big(f_i(\bxstar) - F(\bxstar)\big)^2.
\]}

\begin{theorem}
\label{stationary-point-thm-five}
\stationarypointthmfivetext
\end{theorem}

\paragraph{Discussion.}
We note that the statement of \Cref{stationary-point-thm-five} follows directly from the definition of
$L_\lambda(\bx)$ and is mathematically straightforward. Nevertheless, we state it explicitly to clarify an
important conceptual point: minimizing the variance-regularized objective, $L_\lambda(\bx)$, necessarily leads
to a solution with smaller dispersion in client losses than the minimizer of
the standard average objective, $F(\bx)$.
This result formalizes the intuition that variance regularization directly enforces performance
balance across clients. As such, \Cref{stationary-point-thm-five} provides a clean and explicit link
between $L_\lambda(\bx)$ and our definition of fairness in \Cref{definition_fairness}, making
the fairness effect of $L_\lambda(\bx)$ transparent and unambiguous.

\begin{proof}
Since $\bxlambdastar \in \argmin_{\bx \in \R^d} L_{\lambda}(\bx)$, then  $L_{\lambda}(\bxlambdastar) \leq L_{\lambda}(\bxstar)$.
 We have 
\begin{equation}\label{stationary-point-eq-thirteen}
    {L}_{\lambda}(\bxlambdastar)={F}(\bxlambdastar)+ \avgcnn{\lambda} \big({f}_i(\bxlambdastar) - {F}(\bxlambdastar) \big)^2\leq F(\bxstar) +\avgcnn{\lambda}\big({f}_i(\bxstar) - {F}(\bxstar)  \big)^2.
\end{equation}
This shows that
\begin{equation}\label{stationary-point-eq-fourteen}
    {F}(\bxlambdastar) \leq F(\bxstar) +\avgcnn{\lambda}\big({f}_i(\bxstar) - {F}(\bxstar)  \big)^2 - \avgcnn{\lambda} \big({f}_i(\bxlambdastar) - {F}(\bxlambdastar) \big)^2.
\end{equation}
If $\avgcnn{\lambda}\big({f}_i(\bxstar) - {F}(\bxstar)  \big)^2 - \avgcnn{\lambda} \big({f}_i(\bxlambdastar) - {F}(\bxlambdastar) \big)^2 \geq 0$, then we have 
$$\avgcnn{\lambda} \big({f}_i(\bxlambdastar) - {F}(\bxlambdastar) \big)^2 \leq \avgcnn{\lambda}\big({f}_i(\bxstar) - {F}(\bxstar)  \big)^2.$$ Otherwise, 
if $\avgcnn{\lambda}\big({f}_i(\bxstar) - {F}(\bxstar)  \big)^2 - \avgcnn{\lambda} \big({f}_i(\bxlambdastar) - {F}(\bxlambdastar) \big)^2 < 0$, it means that ${F}(\bxlambdastar) < {F}(\bxstar)$, which is a contradiction with $ \bxstar \in \argmin_{\bx \in \R^d} F(\bx)$. 

\end{proof}

\section{Analysis of the Algorithms}\label{Analysis of Algorithms}

\subsection{Algorithm Descriptions}\label{app:Algorithm:Descriptions}

In this work, we propose and evaluate several methods for achieving performance equitable fairness in FL. Two of these methods incorporate variance-based regularization, focusing on reducing disparities in client losses and gradients, respectively. The third is the q-FFL method, which reweights client contributions based on their local losses to promote uniform performance across clients.
To minimize the objective function in \Cref{eq:fairFL}, which aims to minimize both the average and the variance of client loss functions, we propose two algorithms:
An approximate version (Algorithm~\ref{app:algo:loss}), and an exact version (Algorithm~\ref{algo:exact:lossreall}). Similarly, for the objective in \Cref{eq:fairGrad}, which minimizes the average loss and reduces the variance of the gradients of client loss functions, we propose two algorithms: An an approximate version (Algorithm~\ref{sec:app:algo:grad}), and an
exact version (Algorithm~\ref{sec:app:algo:gradreall}).

In the loss-alignment approximate algorithm (referred to as FairLoss; see  Algorithm~\ref{app:algo:loss}), at iteration~$t$, the server sends the model parameters, $\bxt$, along with the average of client loss functions, $a^t$, and the average client gradient, $\bg^t$, from the previous iteration (iteration $t-1$) to all clients. Each client computes its local loss, $f_i(\bxt)$, and constructs a regularized approximate objective with a penalty term, such that $f_i(\bxt) + \frac{\lambda }{2}(f_i(\bxt) - a^t)^2$ using its own data. The model parameters are updated using one SGD step, $\bxtt_i = \bxt - \eta\cdot \bg_i^t$, where $ \bg_i^t$ is defined based on the previous iteration's information as $\bg_i^{t}= \nabla f_i(\bxt) + \lambda (f_i(\bxt) - a^t)(\nabla f_i(\bxt) - \bg^{t})$. This method reduces communication frequency between server and clients  because it eliminates the need for clients to send their current loss values back to the server and wait for the aggregated average before proceeding with updates. By relying on the previous iteration's aggregated information, clients can perform local updates independently, thereby decreasing the number of communication rounds required per iteration.

In the loss-alignment exact algorithm (referred to as FairLoss$^*$; see Algorithm~\ref{algo:exact:lossreall}), each client first receives the current model parameters $\bxt$ from the server, then computes its local loss and gradient, and sends these values back to the server. The server aggregates the received losses and gradients to compute the current average loss, $a^t$, and average gradient, $\bg^t$, and then broadcasts this information back to the clients. Each client uses the received average loss to construct the regularized exact objective $f_i(\bxt) + \frac{\lambda}{2} (f_i(\bxt) - a^t)^2$.
This objective integrates the global loss information from the current iteration with the client’s local data.
Subsequently, the clients update their model parameters by performing a SGD step using $\bg_i^{t}= \nabla f_i(\bxt) + \lambda (f_i(\bxt) - a^t)( \nabla f_i(\bxt) - \bg^{t}  )$.  While this algorithm provides the best available information for the updates, it requires more communication between clients and the server.

In the gradient-alignment algorithms (FairGrad and FairGrad$^*$; see Algorithms~\ref{sec:app:algo:grad}, \ref{sec:app:algo:gradreall}), a process similar to loss alignment is followed, but the regularization is applied to client gradients instead of losses. In the approximate version (FairGrad), each client uses the average of client gradients from the previous iteration, whereas in the exact version (FairGrad$^*$), the average gradients are computed in the current iteration and communicated back to the clients.

\subsection{Theoretical Analysis of the Algorithms}
\label{app:Theoretical:Analysis:of:Algorithms}

We now prove Theorem~\ref{approximat-loss-thm-alg-one}. We recall the theorem:

\medskip
\noindent\textbf{Theorem~\ref{approximat-loss-thm-alg-one}.}
\textit{\approximatlossthmalgonetext}
\begin{proof}
From Algorithm~\ref{app:algo:loss}, we have $a_i^t=f_i(\bxt) +\frac{\lambda }{2}(f_i(\bxt) - a^t)^2$, and $\bg_i^t=\nabla f_i(\bxt) +{\lambda }(f_i(\bxt) - a^t)(\nabla f_i(\bxt) - \bg^t)$ where $a^t = \avgn {f_i(\bx^{t-1})}= F(\bx^{t-1})$ and $\bg^t = \avgn {\nabla f_i(\bx^{t-1})}= \nabla F(\bx^{t-1})$. We define
\begin{subequations}\label{eq-lemma}
\begin{align}
\abar_\lambda(\bxt)
&:= F(\bxt) 
+ \avgcnn{\lambda} \big(f_i(\bxt) - a^t \big)^2,
\label{eq-lemma-one} \\
\bgbar_\lambda(\bxt)
&:= \nabla F(\bxt) 
+ \avgcn{\lambda} \big(f_i(\bxt) - a^t \big)\big(\nabla f_i(\bxt) - \bg^t \big).
\label{eq-lemma-two}
\end{align}
\end{subequations}
Now using Equations \eqref{eq-lemma-one} and  \eqref{eq:fairFL},
\begin{align*}
{L}_\lambda(\bxt) - \abar_\lambda(\bxt)
    &= F(\bxt) 
    +  \avgcnn{\lambda} \big(f_i(\bxt) - F(\bxt) \big)^2- F(\bxt) 
    -  \avgcnn{\lambda} \big(f_i(\bxt) - a^t  \big)^2\\  
&= \avgcnn{\lambda}\big((f_i(\bxt) - F(\bxt))^2 - (f_i(\bxt) - a^t)^2\big)\\
&= \avgcnn{\lambda} f_i(\bxt)^2 -2 f_i(\bxt)F(\bxt) + F(\bxt)^2 - f_i(\bxt)^2 - (a^t)^2 + 2f_i(\bxt)a^t\\
&= \avgcnn{\lambda}-2f_i(\bxt)( F(\bxt) - a^t) + \big( F(\bxt)^2 -(a^t)^2 \big)\\
&= -\avgcn{\lambda\big(F(\bxt) - a^t\big)}f_i(\bxt) + \frac{\lambda}{2}\big( F(\bxt)^2 -(a^t)^2 \big)\\
&= -\lambda\big(F(\bxt) - a^t\big)F(\bxt) + \frac{\lambda}{2}\big( F(\bxt)^2 -(a^t)^2 \big)\\
&= -\frac{\lambda}{2}F(\bxt)^2 + {\lambda}a^t F(\bxt) - \frac{\lambda}{2}(a^t)^2\\
&=- \frac{\lambda}{2}\big(F(\bxt) - a^t \big)^2\\
&=- \frac{\lambda}{2}\big(F(\bxt) - F(\bx^{t-1}) \big)^2 \leq 0.
\end{align*} 
Now we show that
\[
    \nabla L_\lambda(\xt)- \bgbar_{\lambda}(\bxt)
    = -\lambda\big(\nabla F(\xt)- \nabla F(\x^{t-1})\big)\big(F(\xt)-F(\x^{t-1})\big).
\]
Using the gradients of Equations~\eqref{eq:fairFL} and \eqref{eq-lemma-two}, we have
\begin{align}
\nabla L_{\lambda}(\xt)- \bgbar_{\lambda}(\bxt)
    &= \nabla F(\bxt) 
    +  \avgcn{\lambda} \big(f_i(\bxt) - F(\bxt) \big)\big(\nabla f_i(\bxt) - \nabla F(\bxt) \big) \nonumber\\
    &\qquad\qquad - \nabla F(\bxt) 
    -  \avgcn{\lambda} \big(f_i(\bxt) - a^t \big)\big(\nabla f_i(\bxt) - \bg^t  \big)\nonumber \\  
    &=  \avgcn{\lambda} \big(f_i(\bxt) - F(\bxt) \big)\big(\nabla f_i(\bxt) - \nabla F(\bxt) \big) \nonumber\\
    &\qquad\qquad - \avgcn{\lambda} \big(f_i(\bxt) - a^t \big)\big(\nabla f_i(\bxt) - \bg^t \big)\nonumber \\ 
    &= \avgcn{\lambda} \Big( -f_i(\bxt)\nabla F(\bxt)- F(\bxt)\nabla f_i(\bxt) + F(\bxt)\nabla F(\bxt) \nonumber\\
    &\qquad\qquad\;\; + f_i(\bxt) \bg^t + a^t\nabla f_i(\bxt) - a^t\bg^t \Big) \nonumber \\
    &= \lambda a^t \big(\nabla F(\bxt)-\bg^t\big) + \lambda F(\bxt)\big(\bg^t - \nabla F(\bxt)\big) \nonumber \\
    &= \lambda \big(\nabla F(\bxt)-\bg^t\big)\big(a^t - F(\bxt)\big)\nonumber \\
    &= -\lambda \big(\nabla F(\bxt)-\nabla F(\bx^{t-1})\big)\big(F(\bxt) - F(\bx^{t-1})\big).
    \label{eq:grad:exact:app}
\end{align}
    Using \Cref{eq:grad:exact:app}, we get that \[ \|\nabla L_{\lambda}(\xt) \| \leq  \| \bgbar_{\lambda}(\bxt) \| + \lambda \|\nabla F(\xt) - \nabla F(\x^{t-1}) \| |F(\xt)-F(\x^{t-1})|.\]
    If each $f_i(\bx)$ is $L$-smooth, then 
    \begin{align*}
        \big\|\nabla L_{\lambda}(\xt) \big\| 
            &\leq \big\| \bgbar_{\lambda}(\bxt) \big\|
            + \lambda L \|\xt - \x^{t-1} \| \big|F(\xt) - F(\x^{t-1})\big| \nonumber\\
            &\leq \big\| \bgbar_{\lambda}(\bxt) \big\| + \lambda L \eta\big\| \bgbar_{\lambda}(\x^{t-1}) \big\| \big|F(\xt) - F(\x^{t-1})\big|.
    \end{align*}
    We obtain the last equation from the SGD step, $\xt = \x^{t-1} - \eta\bgbar_{\lambda}(\x^{t-1})$.

Now we prove it for $\abar_{\gamma}(\bxt), \bgbar_{\gamma}(\bxt)$. From Algorithm~\ref{sec:app:algo:grad}, each client computes $f_i(\bxt) +\frac{\gamma }{2}\|\nabla f_i(\bxt) - \bg^t \|^2$, where $\bg^t = \avgn {\nabla f_i(\bx^{t-1})}= \nabla F(\bx^{t-1})$.
We define
\begin{equation}\label{eq-lemma-one_gr}
   \abar_\gamma(\bxt)
    := F(\bxt) 
    +  \avgcnn{\gamma} \big\|\nabla f_i(\bxt) - \bg^t \big\|^2.
\end{equation}
Now using \Cref{eq-lemma-one_gr}, and the fact that for vectors $a, b, c$, we have $\|a - c\|^2 - \|a - b\|^2 = 2a^\top (b - c) + \|c\|^2 - \|b\|^2$, we can write
\begin{align}
    {J}_\gamma(\bxt) -\abar_\gamma(\bxt)
        &= F(\bxt) 
           + \avgcnn{\gamma}\big\|\nabla f_i(\bxt) - \nabla F(\bxt) \big\|^2
        -  F(\bxt) -  \avgcnn{\gamma }\big\|\nabla f_i(\bxt) - \bg^t \big\|^2\nonumber\\  
        &= \avgcnn{\gamma}\big\|\nabla f_i(\bxt) - \nabla F(\bxt) \big\|^2 - \big\|\nabla f_i(\bxt) - \bg^t \big\|^2\nonumber\\
        &= \avgcnn{\gamma} \big\|\nabla F(\bxt) \big\|^2 - \|\bg^t \|^2 + 2\nabla f_i(\bxt)^T\big(\bg^t - \nabla F(\bxt)\big)\nonumber\\
        &= \frac{\gamma }{2} \big\|\nabla F(\bxt) \big\|^2 - \frac{\gamma }{2}\|\bg^t\|^2 + \avgcnn{\gamma} 2\nabla f_i(\bxt)^T\big(\bg^t - \nabla F(\bxt)\big)\nonumber\\
        &= \frac{\gamma}{2} \big\|\nabla F(\bxt) \big\|^2 - \frac{\gamma }{2}\|\bg^t\|^2 + {\gamma}\nabla F(\bxt)^T\bg^t - \gamma\big\|\nabla F(\bxt)\big\|^2\nonumber\\
        &= -\frac{\gamma}{2}\big\|\nabla F(\bxt)\big\|^2 - \frac{\gamma }{2}\|\bg^t \|^2 + {\gamma}\nabla F(\bxt)^T\bg^t\nonumber \\
        &= -\frac{\gamma}{2}\big\|\nabla F(\bxt) - \bg^t\big\|^2 \nonumber \\
        &= -\frac{\gamma}{2}\big\|\nabla F(\bxt) - \nabla F(\bx^{t-1})\big\|^2 \leq 0.
        \label{eq:grad:grad:exact:app}
\end{align} 
Using \Cref{eq:grad:grad:exact:app}, we get that \[\|\nabla J_{\gamma}(\xt) \| \leq  \| \bgbar_{\gamma}(\bxt)  \| + \gamma  \|\nabla F(\xt)- \nabla F(\x^{t-1}) \| \|\nabla^2 F(\xt)\|.\]
If each $f_i(\bx)$ is $L$-smooth, then
\begin{align*}
    \big\|\nabla J_{\gamma}(\xt) \big\| 
        \leq \big\| \bgbar_{\gamma}(\bxt) \big\| 
        + \gamma L^2 \|\xt - \x^{t-1} \|  
        \leq \big\| \bgbar_{\gamma}(\bxt) \big\| +  \gamma L^2 \eta \big\| \bgbar_{\gamma}(\bx^{t-1}) \big\|.
\end{align*}
We obtain the last equation from the SGD step, $\xt = \x^{t-1} - \eta\bgbar_{\gamma}(\x^{t-1})$.
\end{proof}

\subsection{Exact Algorithms}\label{app:exact:algorithms}

In this section, we present the exact versions of the algorithms: FairLoss$^*$ in Algorithm~\ref{algo:exact:lossreall}, FairGrad$^*$ in Algorithm~\ref{sec:app:algo:gradreall}, and the q-FFL in Algorithm~\ref{algo:qffl:qffl} as introduced in \cite{li2019fair}. We included AFL and AAggFF in our experiments by adapting the cross-silo version from \cite{hahn2024pursuing}, using their official implementation as a reference.

\begin{figure}[t]
\centering

% ================= LEFT =================
\begin{minipage}[t]{0.49\textwidth}
\refstepcounter{figure}
\label{algo:exact:lossreall}

\vspace{0pt}
\hrule
\vspace{0.3em}
\textbf{Algorithm \thefigure: FairLoss$^*$}
\vspace{0.3em}
\hrule
\vspace{0.5em}

\begin{algorithmic}
\vspace{-0.5em}
\STATE\textbf{Input:} $\bx^0$, $ \lambda, \eta, T$
\FOR {round $t = 0,1,\ldots,T-1$}
\STATE--- \textbf{Client}-level local training -----------
\vspace{-1.2em}
\FOR {client $i = 1,\ldots,n$}
\STATE Receives $\bx^{t}$ from the server
\STATE Sends $\nabla f_i(\bxt)$ and $f_i(\bxt)$ to the server
\ENDFOR
\STATE--- \textbf{Server}-level aggregation ------------
\vspace{-1.2em}
\STATE Receives all $\nabla f_i(\bxt)$ and $f_i(\bxt)$
\STATE $a^t=\avgn f_i(\bxt)$
\STATE $\bg^t=\avgn \nabla f_i(\bxt)$
\STATE Sends $a^t$ and $\bg^t$ to clients
\STATE--- \textbf{Client}-level local training -----------
\vspace{-1.2em}
\FOR {client $i = 1,\ldots,n$}
\STATE Receives $a^t$ and $\bg^t$ from server
\STATE $a_i^t=f_i(\bxt) +\frac{\lambda }{2}\big(f_i(\bxt) - a^{t} \big)^2$
\STATE $\bg_i^t= \nabla f_i(\bxt)$
\STATE\hspace{2.3em} $+\hspace{0.1em} \lambda \big(f_i(\bxt) - a^t \big)\big( \nabla f_i(\bxt) - \bg^t \big)$
\vspace{-1.2em}
\STATE $\bx_{i}^{t+1} = \bxt - \eta \cdot \bg_i^t$
\STATE Sends $\bx_{i}^{t+1}$ to the server
\ENDFOR
\STATE--- \textbf{Server}-level aggregation ------------
\vspace{-1.2em}
\STATE $\bx^{t+1} = \avgn \bx_{i}^{t+1}$
\STATE Sends $\bx^{t+1}$ to the clients
\ENDFOR
\end{algorithmic}

\vspace{0.3em}
\hrule
\end{minipage}
\hfill
% ================= RIGHT =================
\begin{minipage}[t]{0.49\textwidth}
\refstepcounter{figure}
\label{sec:app:algo:gradreall}

\vspace{0pt}
\hrule
\vspace{0.3em}
\textbf{Algorithm \thefigure: FairGrad$^*$}
\vspace{0.3em}
\hrule
\vspace{0.5em}

\begin{algorithmic}
\vspace{-0.5em}
\STATE\textbf{Input:} $\bx^0, \gamma, \eta, T$
\FOR {round $t = 0,1,\ldots,T-1$}
\STATE--- \textbf{Client}-level local training -----------
\vspace{-1.2em}
\FOR {client $i = 1,\ldots,n$}
\STATE Receives $\bx^{t}$ from the server
\STATE Sends $\nabla f_i(\bxt)$ to the server
\ENDFOR
\STATE--- \textbf{Server}-level aggregation ------------
\vspace{-1.2em}
\STATE Receives all $\nabla f_i(\bxt)$ from the clients
\STATE $\bg^t=\avgn \nabla f_i(\bxt)$
\STATE Sends $\bg^t$ to the clients
\STATE--- \textbf{Client}-level local training -----------
\vspace{-1.2em}
\FOR {client $i = 1,\ldots,n$}
\STATE $\bg_i^t = \nabla \big( f_i(\bxt)$
\STATE\hspace{2.3em} $ + \frac{\gamma}{2} \left\| \nabla f_i(\bxt) - \bg^t \right\|^2 \big)$
\STATE $\bx_i^{t+1} = \bxt - \eta \cdot \bg_i^t$,
\STATE Sends $\bx_{i}^{t+1}$ to the server
\ENDFOR
\STATE--- \textbf{Server}-level aggregation ------------
\vspace{-1.2em}
\STATE Receives all $\bx_{i}^{t+1}$ from the clients
\STATE $\bx^{t+1} = \avgn \bx_{i}^{t+1}$
\STATE Sends $\bx^{t+1}$ to the clients
\ENDFOR
\end{algorithmic}

\vspace{0.3em}
\hrule
\end{minipage}

% ================= LEFT =================
\begin{minipage}[t]{0.49\textwidth}
\refstepcounter{figure}
\label{algo:qffl:qffl}

\vspace{2.5pt}
\hrule
\vspace{0.3em}
\textbf{Algorithm \thefigure: \(q\)-FFL}
\vspace{0.3em}
\hrule
\vspace{0.5em}

\begin{algorithmic}
\vspace{-0.5em}
\STATE\textbf{Input:} $\bx^0, q, \lambda, \eta, T, L=\frac{1}{\eta}$
\FOR {round $t = 0,1,\ldots,T-1$}
    \STATE --- \textbf{Client}-level local training -----------
    \vspace{-1.2em}
    \FOR {client $i = 1,\ldots,n$}
        \STATE Receives $\bx^{t}$ from the server
        \STATE ${\bx}^{t+1}_i = {\bxt} - \eta \cdot\nabla H_q(\bxt)$
        
        \STATE $\Delta {\bxit} = {L} ({\bxt} - {\bx}^{t+1}_i)$
        \STATE $\Delta_i^t =  {f}^q_{i}(\bxt) \Delta \bx_i^t$
        \STATE $h_i^t = q {f}^{q-1}_{i}(\bxt) \|\Delta \bxt_i\|^2 +L {f}^q_{i}(\bxt)$
        \STATE Sends $\Delta_i^t$ and $h_i^t$ to the server
    \ENDFOR
    \STATE --- \textbf{Server}-level aggregation ------------
    \vspace{-1.2em}
    \STATE Receives all $\Delta_i^t$ and $h_i^t$ from the clients
    \vspace{-0.7em}
    \STATE $\bx^{t+1} = \bxt - \frac{\sum_{i=1}^{n} {\Delta_i^t}}{\sum_{i=1}^{n} h_i^t}$ 
    \STATE Sends $\bx^{t+1}$ to the clients
\ENDFOR
\end{algorithmic}

\vspace{0.3em}
\hrule
\end{minipage}
\hfill

\end{figure}

\section{Experiment Setup}\label{appendix:experiment:setup}

\textbf{Datasets}
We evaluated the investigated methods on four standard image datasets, which were MNIST~\cite{lecun1998gradient}, CIFAR-10~\cite{krizhevsky2009cifar}, CIFAR-100~\cite{krizhevsky2009cifar}, and Tiny ImageNet~\cite{le2015tiny}.
For MNIST, we selected a subset of fewer than $1000$ images and distributed them across $50$ clients to simulate data scarcity. A multinomial regression model was used.
For CIFAR-10, we used the full dataset distributed across $50$ clients, training a CNN with two convolutional layers. 
For CIFAR-100 and Tiny ImageNet, we used $50\%$ of each dataset, distributed across $100$ clients, with MobileNetV2 as the model architecture.

In all settings, client datasets were generated using a Dirichlet distribution with parameters $\alpha = 0.05$, $0.1$, and $0.5$, where smaller $\alpha$ values indicate more extreme heterogeneity (arising from both label and quantity skew). Each client maintained their own training, validation, and test datasets.
For each $\alpha$, we ran experiments with $5$ different random initialization seeds. Results were averaged across runs at each epoch to ensure robustness, and we report the average test accuracy together with the variance of test accuracies across clients. Standard errors, shown in parentheses, were computed over the different initializations (see  \Cref{tab:Dirichlet-distribution:cifar100},  \Cref{tab:Dirichlet-distribution:Tiny ImageNet}, \Cref{tab:Dirichlet-distribution:cifar10},  and \Cref{tab:Dirichlet-distribution:mnist}).

\noindent\textbf{Hyperparameter search:}
We performed hyperparameter tuning for all methods using a validation set, exploring a range of learning rates, batch sizes, regularization coefficients ($\lambda$, $\gamma$, and $q$), and the number of global epochs. 
The best hyperparameter configuration for each method was selected based on validation performance and used consistently in all reported experiments.
Specifically, we aimed to balance accuracy and stability by selecting the configuration that maximized the criterion
\[
    \overline{\text{acc}} - t \cdot \sqrt{\frac{ \overline{\text{var}}}{m}},
\]
where $t=1.96$ indicating a $95\%$ confidence interval around the mean, with $\overline{\text{acc}}$ denoting the mean, over seeds, of the average validation accuracy across clients for each seed, $\overline{\text{var}}$ denoting the mean, over seeds, of the sample variance of validation accuracies across clients for each seed, and $m$ is the number of independent experiments (\emph{i.e.}, seeds). This selection strategy aligns with our theoretical fairness definition in \Cref{definition_fairness} by restricting consideration to high-performing configurations and, among these, choosing those with the most equitable performance across clients (\emph{i.e.}, lowest variance).
The learning rates used in the search were $[0.0001,$ $0.001,$ $0.01,$ $0.1,$ $1.0].$
The fairness methods' parameters, such as $\lambda$, $\gamma$, and $q$ were
$[0.0001,$ $0.0003,$ $0.001,$ $0.003,$ $0.01,$ $0.03,$ $0.1,$ $0.3,$ $1.0,$ $3.0,$ $10.0]$, and experiments were conducted for global epochs of $500$, $1000$, and $2000$.
All experiments were run on NVIDIA GPUs: A100 40GB for CIFAR-100 and Tiny ImageNet, and T4 16GB for MNIST and CIFAR-10.

\end{document}